\documentclass[sigconf]{acmart} 

\AtBeginDocument{%
  }

\setcopyright{acmlicensed}
\copyrightyear{2018}
\acmYear{2018}
\acmDOI{XXXXXXX.XXXXXXX}
\usepackage{multirow}
\usepackage{amsmath}
\usepackage{makecell}
\usepackage{graphicx}
\usepackage{makecell}
\usepackage{colortbl}
\usepackage{xcolor}
\usepackage{epstopdf}
\usepackage{subcaption}
\usepackage{float}
\usepackage{algorithm}
\usepackage{algorithmic}
\acmConference[Conference acronym 'XX]{Make sure to enter the correct
  conference title from your rights confirmation emai}{June 03--05,
  2018}{Woodstock, NY}
\acmISBN{978-1-4503-XXXX-X/18/06}

\usepackage[utf8]{inputenc} 
\usepackage[T1]{fontenc}    
\usepackage{hyperref}       
\usepackage{url}            
\usepackage{booktabs}       
\usepackage{amsfonts}       
\usepackage{nicefrac}       
\usepackage{microtype}      
\usepackage{xcolor}         
\usepackage{xspace}
\usepackage{multirow}
\usepackage{graphicx}
\newcommand{\system}{\texttt{SRTFD}\xspace}

\newcommand{\compact}{\vspace{-7pt}}
\newcommand{\subcompact}{\vspace{-5pt}}

\usepackage[skip=8pt]{caption}
\newenvironment{myitemize}
{ \begin{itemize}
		\vspace{-1ex}	
		\setlength{\itemsep}{0pt}
		\setlength{\parskip}{0pt}
		\setlength{\parsep}{0pt}    }
	{ 	 \end{itemize}                    }

\begin{document}

\title{SRTFD: Scalable Real-Time Fault Diagnosis through Online Continual Learning}

\author[1,2]{Dandan Zhao}
\affiliation{%
  \institution{Chongqing University}
  \city{Chongqing}
  \country{China}
}
\affiliation{%
  \institution{Nanyang Technological University}
  \country{Singapore}
}
\email{whsmhgy@gmail.com}

\author{Karthick Sharma}
\affiliation{%
 \institution{University of Sri Jayewardenepura}
 \city{Doimukh}
 \state{Arunachal Pradesh}
 \country{Sri Lanka}}
 \email{en93899@sjp.ac.lk}
 
\author{Hongpeng Yin}
\affiliation{%
 \institution{Chongqing University}
  \city{Chongqing}
  \country{China}
}
\email{yinhongpeng@cqu.edu.cn}

\author{Yuxin Qi}
\affiliation{%
 \institution{Shanghai Jiao Tong University}
 \city{Shanghai}
  \state{Beijing Shi}
 \country{China}}
\email{qiyuxin98@sjtu.edu.cn}

\author{Shuhao Zhang}
\affiliation{%
 \institution{Nanyang Technological University}
 \country{Singapore}}
\email{shuhao.zhang@ntu.edu.sg}


\begin{abstract}

Fault diagnosis (FD) is essential for maintaining operational safety and minimizing economic losses by detecting system abnormalities. Recently, deep learning (DL)-driven FD methods have gained prominence, offering significant improvements in precision and adaptability through the utilization of extensive datasets and advanced DL models. Modern industrial environments, however, demand FD methods that can handle new fault types, dynamic conditions, large-scale data, and provide real-time responses with minimal prior information. Although online continual learning (OCL) demonstrates potential in addressing these requirements by enabling DL models to continuously learn from streaming data, it faces challenges such as data redundancy, imbalance, and limited labeled data. To overcome these limitations, we propose \system, a scalable real-time fault diagnosis framework that enhances OCL with three critical methods: \textit{Retrospect Coreset Selection (RCS)}, which selects the most relevant data to reduce redundant training and improve efficiency; \textit{Global Balance Technique (GBT)}, which ensures balanced coreset selection and robust model performance; and \textit{Confidence and Uncertainty-driven Pseudo-label Learning (CUPL)}, which updates the model using unlabeled data for continuous adaptation. Extensive experiments on a real-world dataset and two public simulated datasets demonstrate \system's effectiveness and potential for providing advanced, scalable, and precise fault diagnosis in modern industrial systems.
\end{abstract}





\maketitle

\section{Introduction} 
Faults in engineering systems pose substantial risks to both performance and safety~\cite{Renyw}. For instance, faults in steel hot rolling processes can lead to defective steel slabs, resulting in significant scrap losses and compromising worker safety. Similarly, faults in autonomous vehicles can cause navigation errors or system malfunctions, potentially leading to accidents and endangering passengers and pedestrians. Consequently, fault diagnosis (FD) – the process of detecting and identifying potential faults within these systems – is crucial for maintaining reliability, safety, and economic efficiency~\cite{Abid}. While recent research in computer vision has focused on industrial anomaly detection (IAD)~\cite{Alex, Aimira, WangCJ, FangZ, Fedor}, primarily addressing quality faults in end products using image data to identify statistical features and pattern changes, FD targets process faults. FD is essential for immediate fault detection and intervention, focusing on specific functional problems using sensor data to prevent system failures.

\begin{table}[t]
\centering
\caption{The existing state-of-the-art FD methods and the key challenges in modern FD tasks.}\label{Existing_FD}
\resizebox{\linewidth}{!}{
\begin{tabular}{|c|c|c|c|c|c|c|}
\hline
 \multicolumn{1}{|c|}{\multirow{2}{*}{Methods}}&\multicolumn{1}{c|}{\multirow{2}{*}{Year}}  & \multicolumn{5}{c|}{\textbf{Key Challenges}} \\ \cline{3-7}
   &  & \textbf{\makecell{Low-training\\ cost}} & \textbf{\makecell{High-data\\ efficiency}} & \textbf{\makecell{Large-scale\\ data}} & \textbf{\makecell{Imbalance\\ data}} & \textbf{\makecell{Limited\\ labeled data}} \\ \hline
  \rowcolor{gray!20}OSELM \cite{Haow} & 2020 & $\times$ & $\times$ & $\times$ & \checkmark & $\times$ \\ \cline{2-7}
 D-CART \cite{Denghx} & 2020 & \checkmark & $\times$ & $\times$ & $\times$ & $\times$ \\ \hline
 \rowcolor{gray!20}TCNN \cite{Xu} & 2020 & \checkmark & $\times$ & $\times$ & $\times$ & $\times$ \\ \hline
 OSSBLS \cite{Puxk} & 2021 & $\times$ & $\times$ & $\times$ & $\times$ & \checkmark \\ \hline
 \rowcolor{gray!20}AMPNet \cite{Fedor} & 2022 & $\times$ & $\times$ & \checkmark & $\times$ & $\times$ \\ \hline
 1D-CNN \cite{Elsisi} & 2022 & $\times$ & $\times$ & $\times$ & $\times$ & \checkmark \\ \hline
 \rowcolor{gray!20}TVOTL \cite{Zhouyx} & 2023 & $\times$ & $\times$ & $\times$ & $\times$ & \checkmark \\ \hline
 ODDFD \cite{Jinlh} & 2023 & $\times$ & $\times$ & $\times$ & $\times$ & \checkmark \\ \hline
 \rowcolor{gray!20}ODCT \cite{Linjh} & 2024 & $\times$ & $\times$ & $\times$ & \checkmark & $\times$ \\ \hline
 OLFA \cite{Yan} & 2024 & $\times$ & $\times$ & $\times$ & $\times$ & \checkmark \\ \hline
 \rowcolor{gray!20} MPOS-RVFL \cite{HanPY} & 2024 & \checkmark & $\times$ & $\times$ & \checkmark & \checkmark \\ \hline
 \system (\textbf{Ours}) & 2024 & \checkmark & \checkmark & \checkmark & \checkmark & \checkmark \\ 
\hline
\end{tabular}\label{OLFD-approaches}
}
\end{table}
FD techniques have evolved significantly over time. In the 19th century, FD began with basic limit checking and progressed to model-based FD, which involves physically modeling the entire industrial system's working process and incorporating statistical methods such as trend analysis and parameter estimation~\cite{Gaozw}. The evolution continued with the development of knowledge-based FD, which utilizes signal models and spectral analysis~\cite{Nanc}. The advent of artificial intelligence and neural networks marked a new era of data-driven FD~\cite{Chen, Wu}. Today, the explosion of data has further propelled the development of deep learning (DL)-driven FD methods. By leveraging large datasets and sophisticated DL models, these methods have significantly enhanced the precision and adaptability of FD, making them crucial for modern industrial environments~\cite{Chenxh}. Our work builds on this progression, aiming to address the challenges of real-time fault diagnosis in dynamic and large-scale systems with DL-driven FD methods.

Today, as equipment and systems become increasingly complex, refined, large-scale, and digitalized, the demands for modern FD methods have evolved significantly~\cite{Zhujj}. Modern industrial environments require FD methods that can handle new fault types, adapt to changing conditions, process large-scale data efficiently, provide real-time responses, and operate with minimal prior information~\cite{HanPY}. These requirements present several key technical challenges, including managing training costs, ensuring data efficiency, processing large-scale data, handling data imbalance, and working with limited labeled data. Furthermore, the continuous and dynamic nature of industrial processes necessitates handling streaming data and maintaining model performance over time. Existing DL-based FD methods often struggle with these challenges, as summarized in Table~\ref{Existing_FD}. They particularly face issues related to high computational costs, inefficiencies in handling streaming data, and difficulties in sustaining model performance due to data redundancy and imbalance. This underscores the necessity for more advanced and integrated FD techniques that can meet the evolving needs of modern industrial systems.

Online continual learning (OCL)~\cite{Ghunaim} offers a promising solution to the limitations of traditional DL-based FD methods by enabling models to continuously learn from streaming data. This capability is crucial for adapting to new fault types, changing operating conditions, and meeting real-time requirements. Recent advancements in OCL, such as CAMEL~\cite{Li} and GoodCore~\cite{Chai}, have focused on reducing training costs and improving data efficiency. However, directly applying these OCL approaches to fault diagnosis is not straightforward due to domain-specific challenges including 1) highly redundant system monitoring data, 2) extreme data imbalance, and 3) the difficulty of collecting sufficient labeled samples. Therefore, while OCL provides a robust framework, it requires significant adaptation to effectively meet the unique demands of real-time fault diagnosis in modern industrial environments.

To address the limitations of traditional and OCL-based fault diagnosis methods, we propose \system, a scalable real-time fault diagnosis framework composed of three key components: Retrospect Coreset Selection (RCS), Global Balance Technique (GBT), and Confidence and Uncertainty-driven Pseudo-label Learning (CUPL). RCS enhances data efficiency and reduces training costs by selecting the most relevant data for model updates, thus avoiding redundant training on unnecessary monitoring data. GBT tackles data imbalance by ensuring balanced coreset selection and maintaining robust model performance. CUPL enables model updates using unlabeled data, addressing the scarcity of labeled monitoring data and facilitating continuous adaptation. By integrating these components, \system effectively handles the complexities of real-time fault diagnosis in modern industrial environments, characterized by large-scale, streaming, and unbalanced data with minimal labeled samples.

We validated the effectiveness of \system through experiments on one real-world industrial process (Hot Roll of Steel, HRS) and two simulation processes (Tennessee Eastman Process, TEP, and CAR Learning to Act, CARLA). The HRS dataset includes motor current data from 294 rollers across five fault categories, characterized by high imbalance and limited samples. The TEP dataset has 21 fault categories, and the CARLA dataset includes single-sensor and multi-sensor faults under three weather conditions, simulating real-world variability. We compared \system against five benchmarks: a baseline experience replay (ER) model~\cite{ChaudhryA}, CAMEL~\cite{Li}, ASER~\cite{Shimd}, AGEM~\cite{Lopez}, and MPOS-RVFL~\cite{HanPY}, focusing on class-incremental learning and varying working conditions. Our approach showed superior performance and efficiency, with improvements of 2.73\% in recall, 1.43\% in precision, 1.76\% in F1 score, 1.81\% in G-means, and a 55.83\% reduction in training time compared to state-of-the-art FD and OCL methods. These results underscore the effectiveness and cost-efficiency of \system for industrial applications.

The main contributions of this work are:
\begin{myitemize}
    \item We propose \system\footnote{Our code and public datasets are available at: \url{https://anonymous.4open.science/status/SRTFD-F813}}, a novel and scalable real-time fault diagnosis framework designed to enhance online continual learning (OCL) for industrial applications. This framework effectively addresses the limitations of existing fault diagnosis methods, particularly in handling large-scale, streaming, and imbalanced data with minimal labeled samples.
    \item \system integrates three innovative components: Retrospect Coreset Selection (RCS) to enhance data efficiency and reduce training costs, Global Balance Technique (GBT) to tackle data imbalance and maintain robust model performance, and Confidence and Uncertainty-driven Pseudo-label Learning (CUPL) to enable continuous model updates using unlabeled data.
    \item We validate \system through extensive experiments on real-world and simulated datasets, demonstrating superior performance and efficiency compared to state-of-the-art methods. Our approach achieved notable improvements in recall, precision, F1 score, G-means, and training time, underscoring its effectiveness and cost-efficiency for real-time fault diagnosis in modern industrial environments.
\end{myitemize}
\compact
\section{Background and motivation}

\subsection{Traditional Data-driven-based FD}
Traditional data-driven FD involves three steps: data collection, model training, and fault prediction. Let \( X^{tr} = \{x_1^{tr}, x_2^{tr}, \ldots, x_{n_{tr}}^{tr}\} \) and \( X^{te} = \{x_1^{te}, x_2^{te}, \ldots, x_{n_{te}}^{te}\} \) be the training and testing samples, respectively. Corresponding labels are \( Y^{tr} = \{y_1^{tr}, y_2^{tr}, \ldots, y_{n_{tr}}^{tr}\} \) and \( Y^{te} = \{y_1^{te}, y_2^{te}, \ldots, y_{n_{te}}^{te}\} \), with \( n_{tr} \) and \( n_{te} \) being the number of samples. Labels range from 0 to \( c \), where 0 indicates normal samples and 1 to \( c \) denote different fault categories, with \( c \) being the total number of fault categories.

Features of collected samples are extracted by \( \phi(\cdot) \) before model training. The FD model is then trained using the following loss function:
\vspace{-4pt}
\begin{equation}
\label{E_loss}
\text{min} \sum_{i=1}^{n_{tr}} \mathcal{L}(f(\phi(x_i^{tr}); \theta), y_i^{tr}),
\end{equation}
where, \( \theta \) denotes the model parameters, \( x_i^{tr} \) is the \(i\)-th training sample (\(i =1, 2, \dots, n_{tr}\)). For a new test sample \( x_j^{te} \) (\(j =1, 2, \dots, n_{te}\)), the corresponding label \( y_j^{te} \) can be predicted by the trained FD model as follows:
\vspace{-4pt}
\begin{equation}
\label{E_test}
    \hat{y}_j^{te} = f(\phi(x_j^{te}); \theta).
\end{equation}

Equations (\ref{E_loss}) and (\ref{E_test}) show that fault diagnosis performance depends on the dataset quality. As monitoring data evolves, the model's performance degrades, necessitating data recollection and model retraining. This highlights a key limitation of traditional fault diagnosis methods in meeting real-world demands.

\subsection{Online Continuous Learning}
OCL~\cite{Ghunaim} may address several limitations of traditional FD by enabling models to learn incrementally from streaming data. In OCL, monitoring data is collected over time as \(X^t \in \mathbb{R}^{d\times n} = \{x_1^t, x_2^t, \ldots, x_n^t\}\) and \(Y^t \in \mathbb{R}^{1\times n} = \{y_1^t, y_2^t, \ldots, y_n^t\}\), where \(d\) is the sample dimension, \(n\) is the number of samples, and \(t\) is the collection time. Thus, the loss function equation (\ref{E_loss}) becomes:
\vspace{-4pt}
\begin{equation}
\label{E_Loss_OCL}
\text{min} \sum_{i=1}^{n} \mathcal{L}(f(\phi(x_i^t); \theta_t), y_i^t),
\end{equation}
where, \( x_i^t \in X^t \), \( y_i^t \in Y^t \), and \(\theta_t\) are updated continuously as new data arrives. To update the model at time \( t \) without losing previous information, \(\theta_t\) is adjusted based on \(\theta_{t-1}\) and the loss function gradient. For models optimized by stochastic gradient descent (SGD)~\cite{Ruder}, the update rule is:
\vspace{-4pt}
\begin{equation}
\label{E_Model_update_OCL}
    \theta_t \leftarrow \theta_{t-1} - \eta \frac{1}{n} \sum_{i=1}^n w^t_i \nabla \mathcal{L}(f(\phi(x_i^t); \theta_{t-1}), y_i^t),
\end{equation}
where, \(\eta\) is the learning rate and \(w_i\) is the weight of data \((x_i^t, y_i^t)\). Typically, \(w^t_i = 1\) for all \(i \in [n]\), assuming all samples are equally important. This continuous learning process allows the model to adapt to new fault classes and variable working conditions without retraining from scratch.

Although OCL addresses evolving monitoring data, it has a critical drawback: high model update costs when the data volume at time \(t\) is large. \textit{Coreset selection}~\cite{Li} mitigates this by selecting a smaller representative subset from the current batch, reducing data volume and computational requirements. This approach lowers training complexity, allowing fault diagnosis systems to efficiently manage large datasets and enhancing the scalability and performance of OCL models.

\subsection{Motivation}
Developing a robust FD framework using OCL and coreset selection is promising but faces challenges due to the unique characteristics of industrial monitoring data.

\begin{myitemize}
\item \textbf{Redundancy:} Monitoring data is highly redundant, making OCL methods inefficient for model updates. Coreset selection within each batch overlooks global information, leading to unnecessary updates and extra training costs when consecutive batches are similar.

\item \textbf{Data Imbalance:} FD data is imbalanced, with more normal operating data than fault data, and varying frequencies of different faults. This imbalance results in an unbalanced coreset, reducing FD system performance and reliability.

\item \textbf{Labeled Data Scarcity:} Collecting labeled monitoring data is challenging due to the time-consuming and labor-intensive nature of manual labeling. The lack of labeled data makes model updating difficult.
\end{myitemize}

To address these challenges, we propose \system, a scalable real-time fault diagnosis method through OCL. The details of this approach will be introduced in the next section.
\compact
\section{Problem Statement}
Achieving a realistic fault diagnosis framework in complex systems is challenging due to several key issues mentioned in the previous section. These problems are formulated in this section. 

Let \(D^0 = (X^0, Y^0)\) represent the data used for model pre-training, and \(X_u^t\) denote the arriving unlabeled data. The pseudo-labels \(Y_u^t\) for \(X_u^t\) can be predicted by well-trained model \(f_{\theta_t}(\cdot)\). To handle the large-scale data that accumulates over time \(t\), a small subset \(S^t = (Z^t, V^t)\) is selected from the pseudo-labeled samples \(U_u^t = (X_u^t, Y_u^t)\). The semi-supervised learning strategy is employed by incorporating a small amount of labeled data \(D^t = (X^t, Y^t)\) to update the model along with the pseudo-labeled samples. Thus, the loss function at time \(t\) becomes:
\vspace{-5pt}
\begin{equation}
    \label{E_Loss_FD_1}
    \begin{aligned}
    \text{min} &\sum_{i=1}^{n} \mathcal{L}(f(\phi(x_{i}^t); \theta_t), y_{i}^t) 
    + \sum_{j=1}^{s} \mathcal{L}(f(\phi(z_{j}^t); \theta_t), v_{j}^t),
    \end{aligned}
\end{equation}
where \(z_{j}^t \in Z^t\), \(v_{j}^t \in V^t\), and \(s\) denotes the total number of samples in the subset \(S^t\). The model parameter update rule becomes:
\vspace{-5pt}
\begin{equation}
\label{E_Model_update_FD_1}
\begin{aligned}
    &\theta_t \leftarrow \theta_{t-1} - \eta (A + B),\\
    A = &\frac{1}{n} \sum_{i=1}^n w^t_i \nabla \mathcal{L}(f(\phi(x_i^t); \theta_{t-1}), y_i^t),\\
    B = &\frac{1}{s} \sum_{j=1}^{s} w^t_{j} \nabla \mathcal{L}(f(\phi(z_{j}^t); \theta_{t-1}), v_{j}^t),
\end{aligned}
\end{equation}
where, \(w^t_i\) and \(w^t_{j}\) are the weights of the labeled and pseudo-labeled samples, respectively, and \(\eta\) is the learning rate. The buffer \(B^t\) is crucial in OCL to avoid catastrophic forgetting. It contains data from the pseudo-labeled dataset \(S^t\) and the labeled dataset \(D^t\), formulated as \(B_t \subseteq (S_u^t \cup D^t)\). This allows the model to review and learn from past data while updating with new information.

The goal of SRTFD is to select an effective coreset from the current batch of data while considering historical data. First, the coreset must effectively represent the current batch of data. Thus, we have:
\vspace{-5pt}
\begin{equation}
    \label{E_Loss_core_1}
    \begin{aligned}
    \min \bigg[ \sum_{j=1}^{s} \mathcal{L}(f(\phi(z_{j}^t); \theta_t), v_{j}^t) &- \sum_{k=1}^{u} \mathcal{L}(f(\phi(x_{u}^t); \theta_t), y_{u}^t) \bigg].
    \end{aligned}
\end{equation}

However, the coreset from the above objective function only considers the current batch. To account for historical data, the selected coreset at each time step should not overlap with previous coresets. For the entire coreset \( S^{T}, T = 1, 2, \dots, t-1\), it should meet the condition \(S^t \cap S^{T} = \varnothing\). Additionally, to address the internal imbalance, the proportion of samples from each class in the coreset should be equal. If there are \(c\) fault categories and the coreset size is \(s\), the ideal probability for each class is \(\frac{c}{s}\). Thus, we have:
\vspace{-5pt}
\begin{equation}
    \label{E_Loss_core_Balanced}
    \begin{aligned}
    \min \lambda &\left( \frac{1}{c} \sum_{l=1}^{c} \left| p_l^t - \frac{c}{s} \right| \right),
    \end{aligned}
\end{equation}
where \(\lambda\) is a regularization parameter that balances the importance of the class proportion term. \(p_l^t\) is the proportion of samples from class $l$ in the selected coreset \(S_u^t\).

Moreover, to handle model updates with insufficient labeled data, a pseudo-labeling strategy can leverage the abundance of unlabeled data. Directly using the model's predicted labels as pseudo-labels is intuitive but can result in mislabeled data, degrading performance. Therefore, generating accurate pseudo-labels is crucial. Let \(v_j^t \leftarrow f(\phi(z_j^t); \theta_{t-1})\) denote the process of generating pseudo-labels. The entire problem of achieving a robust and realistic SRTFD can be formulated as follows:
\vspace{-4pt}
\begin{equation}
    \label{E_Loss_core_al}
    \begin{aligned}
    \min \bigg[ \sum_{j=1}^{s} \mathcal{L}(f(\phi(z_{j}^t); \theta_t), v_{j}^t) &- \sum_{k=1}^{u} \mathcal{L}(f(\phi(x_{u}^t); \theta_t), y_{u}^t) \bigg] \\+\lambda &\left( \frac{1}{c} \sum_{l=1}^{c} \left| p_l^t - \frac{c}{s} \right| \right),\\
    \text{s.t.} \quad S^t \cap S^{T} &= \varnothing, \quad T = 1, 2, \dots, t-1,\\
    v_j^t \leftarrow &f(\phi(z_j^t); \theta_{t-1}).
    \end{aligned}
\end{equation}

\compact

\begin{figure}
    \centering
    \includegraphics[width=0.5\textwidth]{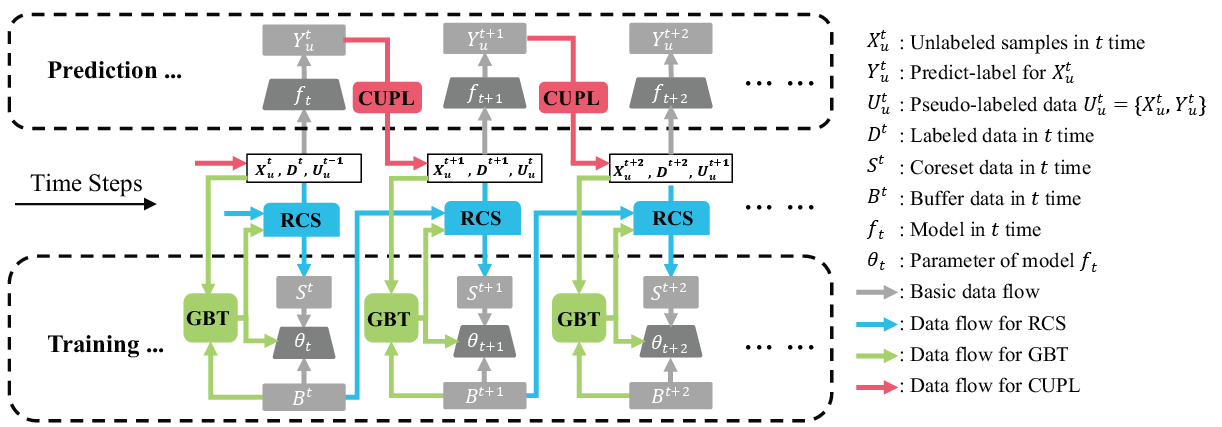} 
    \caption{\system framework conducts fault diagnosis in two stages: Prediction and Training. In the Prediction stage, unlabeled samples are pseudo-labeled and combined with labeled data for model updates via CUPL. In the Training stage, the model is iteratively trained using RSC and GBT to ensure effective learning despite class imbalances.}
    \label{SRTFD}
\end{figure}

\section{Methodology}
This section introduces the \system framework, illustrated in Figure~\ref{SRTFD}. The prediction and training processes are synchronized, with most data for model updating coming from previously unlabeled data and only a small portion being labeled. This approach is realistic for modern complex systems. The framework consists of three main components: Retrospect Coreset Selection (RCS), Global Balance Technique (GBT), and Confidence and Uncertainty-driven Pseudo-label Learning (CUPL). Each component will be detailed in the following subsections.

\subsection{Retrospect Coreset Selection (RCS)}

To optimize fault diagnosis (FD) tasks, we introduce Retrospect Coreset Selection (RCS). RCS selects a representative coreset by considering all historical data, and addressing memory constraints through efficient implementation. Specifically, we use the buffer in the system to store samples that encapsulate all historical data, enabling effective RCS. By assessing the similarity between incoming data and buffer data, we eliminate redundancy. When new data arrives, the buffer filters out redundant information, ensuring that coreset selection focuses on novel, non-redundant data. This approach also reduces computational load by bypassing updates when new batches closely resemble existing data.

Given the large-scale nature of the data, directly computing Euclidean distances between new samples and buffer entries is computationally intensive. To mitigate this, we employ batch-wise metrics and cluster the buffer data. Let \(B^t = \{(x_{b1}^t, y_{b1}^t), \cdots, (x_{bn}^t, y_{bn}^t)\}\) represent the buffer at time \(t\), where \(x_{bi}^t \in (X^t \cup X_u^t)\) and \(y_{bi}^t \in (Y^t \cup Y_u^t)\), with \(i = 1, 2, \cdots, bn\), and \(bn\) being the total number of buffer samples at time \(t\). Labels \(y_{bi}^t\) fall within \(\{0, 1, ..., bc_t\}\), where \(bc_t\) is the number of classes at time \(t\). We partition buffer data into \(bc_t\) clusters based on these labels, denoted as \(\{X_{b1}^t, X_{b2}^t, \ldots, X_{bc_t}^t\}\). By clustering the buffer data, we reduce the computational burden, enabling batch-wise comparison instead of individual distance calculations.


\begin{figure}
    \centering
    \includegraphics[width=0.35\textwidth]{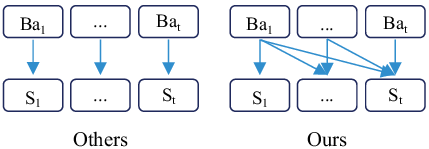} 
    \caption{Comparison of existing~\cite{Li} and our coreset selection approaches. The \(\text{Ba}_\text{t}\) and \(\text{S}_\text{t}\) denote the batch data and selected coreset in t time, respectively.}
    \label{Coreset}
\end{figure}

For the newly arrived data \(X_u^t\), we employ the MiniBatchKMeans algorithm~\cite{Sculleyd}, a highly efficient clustering method designed for large-scale datasets, to partition the data into different clusters. This algorithm accelerates computation and reduces memory usage by processing data in small batches. By clustering \(X_u^t\) into \(uc\) clusters, denoted as \(\{X_{u1}^t, X_{u2}^t, \ldots, X_{uc}^t\}\), where \(X_{ui}^t\) represents the \(ui\)-th cluster and \(ui = 0, 1, \ldots, uc\). The number of clusters \(uc\) must be less than or equal to the number of samples in the arriving batch \(un\), and there is no clustering performed if \(uc\) equals \(un\). Then, the similarity between \(X_{ui}^t\) and \(B_{bi}^t\) can be calculated by Kullback-Leibler (KL) divergence~\cite{Kullback}, which is denoted as:
\vspace{-4pt}
\begin{equation}
    d_{bi,ui}^t = KL(X_{bi}^t, X_{ui}^t).
\end{equation}

If $d_{bi,ui}^t$ is less than a threshold $\tau$, the cluster $X_{ui}^t$ is considered non-redundant. Then, we have
\vspace{-4pt}
\begin{equation}
    S^t = \{\bigcup_{ui=1}^{uc}X_{ui}^t  \mid d_{bi,ui}^t \leq \tau
    \label{filter}
\end{equation}

This buffer retrospective process ensures that only the most informative and non-redundant data points are retained in the coreset. Thus, the goal of \(S_u^t \cap S_u^{T} = \varnothing\), for \(T = 1, 2, \ldots, t-1\), is achieved. 

Selecting the most representative samples is recognized as an NP-hard problem. However, in the recent OCL technique Camel~\cite{Li}. transformed this challenge into a submodular maximization problem. Building on their work, we iteratively select two samples with the largest Euclidean distance until the coreset size requirement is satisfied. The selection of the coreset \(S^t\) in RCS is defined as follows:

\begin{equation}
S^t = \{Z^t \subseteq \bigcup_{ui=1}^{uc} X_{ui}^t \mid d_{bi,ui}^t \geq \tau, d_{\max}\},
\label{core}
\end{equation}
here, \(d_{\max} = \max_{i,j \in [s]} ||z_i - z_j||\), with \(z_i, z_j \in Z^t\) and \(s\) being the size of the coreset. \(Z^t\) is a subset of \(\bigcup_{ui=1}^{uc} X_{ui}^t\). The distance metric \(d_{bi,ui}^t\) ensures that only data points with a similarity measure above a threshold \(\tau\) are selected. \(d_{\max}\) represents the maximum Euclidean distance between any two samples within \(Z^t\).

\subsection{Global Balance Technique (GBT)}
As highlighted in the introduction, the imbalance problem significantly impacts the performance of fault diagnosis systems. Current online continual learning (OCL) methods rarely address this issue adequately. They typically attempt to mitigate it by selecting a coreset that appears balanced from the current batch data, prioritizing samples from less-represented classes. However, this often results in a pseudo-balanced coreset that does not reflect a truly balanced distribution. For instance, if a class has many samples in the buffer but few in the current batch, existing OCL methods will select more samples from this class, neglecting genuinely underrepresented classes. Furthermore, in cases of extreme imbalance, selecting a balanced coreset alone cannot resolve the problem, as the selected coreset often remains unbalanced. Therefore, we propose GBT, which addresses this issue based on two critical factors affecting fault diagnosis performance: the quality of training data and the frequency of model updates.

From the perspective of training data, feeding a balanced dataset into the model will undoubtedly enhance its performance. However, the FD model is trained on both the current batch of data and the data stored in the buffer. Focusing solely on the current batch data makes it easy to obtain a pseudo-balanced coreset. Thus, it is essential to consider both the buffer samples and the categories, and the corresponding objective function (\ref{E_Loss_core_Balanced}) becomes: 
\vspace{-4pt}
\begin{equation}
    \begin{aligned}
    \min &\left( \frac{1}{c} \sum_{l=1}^{c} \left| ps_{l}^t - \frac{c}{s} \right| \right) + &\left( \frac{1}{bc_t} \sum_{l=1}^{bc_t} \left| pb_{l}^t - \frac{bc_t}{bn} \right| \right)
    \label{Imbalance}
    \end{aligned}
\end{equation}
where \( ps_{l}^t \) and \( pb_{l}^t \) represent the true probabilities of the \( l \)-th classes in the current coreset and buffer, respectively. \( bn \) represents the total sample size in the buffer, and \( bc_t \) denotes the number of classes in the buffer at \(t\) time. This approach ensures a genuinely balanced dataset by selecting samples from the least represented classes in both the buffer and the current coreset.

From the perspective of model updates, considering the scenario where certain classes have a significantly low number of samples, with \( ps_l^t \ll \frac{c}{s} \) and \( pb_l^t \ll \frac{bc_t}{b} \), it becomes evident that selecting a balanced coreset is unfeasible. Therefore, we introduce the following loss function:
\vspace{-4pt}
\begin{equation}
    \mathcal{L} = - \sum_{i=1}^{b} (1 - p_i)^\gamma \log(p_i) - \sum_{j=1}^{s} \alpha (1 - p_j)^\gamma\log(p_j)
    \label{Loss_imbalance}
\end{equation}
where \( p_i  = f_{\theta_t}(x_{bi})\) and \( p_j = f_{\theta_t}(z_j)\) are the predicted probabilities for data from the buffer and coreset, respectively. \( \alpha \) represents the data weights in coreset. The modulating factor \(\gamma\) reduces the loss from easy-to-classify examples, encouraging the model to focus on harder cases. Equation (\ref{Loss_imbalance}) directs the model's attention towards underrepresented and difficult classes, effectively mitigating the class imbalance issue.

\subsection{Confidence and Uncertainty-driven Pseudo-label Learning (CUPL)}
The proposed CUPL module enhances the \system training process by utilizing both labeled and unlabeled data. The critical components of this method include generating pseudo-labels and employing a selection strategy to incorporate these labels into the training set. Here, the pseudo-labels \( v_j \) are determined directly by the maximum probability, expressed as \( v_j = \Gamma[\max(p_j)] \). For the selection strategy, the conventional approach for selecting pseudo-labeled samples is to choose those with high confidence. However, since most newly arriving data are normal data, the model shows high confidence in predicting these normal instances but low confidence in identifying rare faults. Thus, relying solely on high-confidence selections can degrade overall performance in FD tasks. 

Inspired by the work of Mamshad et al.~\cite{Rizve}, we introduce both positive and negative label selection. Specifically, samples are selected as positive pseudo-labels when the model's predicted probability is high, and as negative pseudo-labels when the probability is extremely low, which can be denoted as follows:
\vspace{-4pt}
\begin{equation}
    g_j = \mathbf{1}[p_j \geq \tau_p] + \mathbf{1}[p_j \leq \tau_n], 
\end{equation}
where \( g_j \) is an indicator that determines whether the sample \( j \) is selected. The indicator function \( \mathbf{1}[\cdot] \) returns 1 if the condition inside the brackets is true and 0 otherwise. The thresholds \( \tau_p \) and \( \tau_n \) are used to select positive and negative pseudo-labels, respectively. 

Additionally, it has been demonstrated that predictions with low uncertainty are more likely to result in correct pseudo-labels, as shown by Mamshad's research. Therefore, we incorporate an uncertainty constraint into the positive and negative pseudo-label selection process. The uncertainty of model predictions can be estimated using Monte Carlo dropout (MCdropout)~\cite{Malp}. MCdropout estimates uncertainty by incorporating dropout during the inference phase. The process involves performing multiple stochastic forward passes through the network, each time applying dropout with a certain probability \( p \). For a layer with output \( \mathbf{h} \), the output with dropout is \( \mathbf{h}' = \mathbf{h} \odot \mathbf{d} \), where \( \mathbf{d} \sim \text{Bernoulli}(p) \). Repeating this process \( T \) times results in a set of predictions \( \{\mathbf{\hat{y}}^{(1)}, \mathbf{\hat{y}}^{(2)}, \ldots, \mathbf{\hat{y}}^{(T)}\} \). The mean prediction is calculated as \( \mathbf{\hat{y}}_{\text{mean}} = \frac{1}{T} \sum_{t=1}^T \mathbf{\hat{y}}^{(t)} \), and the uncertainty is estimated using the variance \( \mathbf{\hat{\sigma}}^2 = \frac{1}{T} \ \sum_{t=1}^T (\mathbf{\hat{y}}^{(t)} - \mathbf{\hat{y}}_{\text{mean}})^2 \). This approach is simple to implement, requiring no complex modifications to the model architecture, and provides valuable insights into the model's confidence in its predictions, especially with uncertain or complex input data.

By utilizing both the confidence and uncertainty of network predictions, a more accurate subset of pseudo-labels can be selected. The selection function becomes:
\vspace{-4pt}
\begin{equation}
g_i = \mathbf{1}\left[\hat{\sigma}^2 \leq \kappa \text{ and } p_j \geq \tau_p\right] + \mathbf{1}\left[\hat{\sigma}^2 \leq \kappa \text{ and } p_j \leq \tau_n\right],
\label{Pseudo}
\end{equation}
where \( \kappa \) is the uncertainty thresholds. 

\compact
\section{Experiments}
\begin{table*}[t]
  \begin{center}
    \caption{The description of datasets}\label{DataDescription}
    \resizebox{0.7\linewidth}{!}{
    \begin{tabular}{c|cccccc}
    \hline
    \textbf{Dataset}  &\textbf{\makecell{Normal samples}}   & \textbf{\makecell{Number of fault classes}} &\textbf{Number of each fault} & \textbf{Dimension} & \textbf{\makecell{Number of\\ working cond.}} & \textbf{\makecell{Total\\ samples}}\\ \hline
      HRS   &36333   & 5  &1783/8114/1533/83/83 & 120 & 1&47929\\
      TEP   &4320     & 21 &800 & 52  & 1& 31200\\
      \makecell{CARLS (single-sensor)} &89166      & 9  &\makecell{2404/1803/2404/2404\\ /2404/1604/2004/1803/2004}      &10   &3 &108000\\
      \makecell{CARLS (multi-sensor)}  &100885     & 4    & 1604/1604/1503/2404   &10  &3 &108000 \\ \hline
    \end{tabular}}
  \end{center}
\end{table*}
\subsection{Experimental Setup}
Experiments were conducted on an Intel(R) Xeon(R) w7-3455 system with an NVIDIA RTX 6000 Ada Generation GPU (48GB GDDR6) and 512GB of RAM. The software used includes Python 3.12.4 and PyTorch 2.3.0 with CUDA 12.1.

\textbf{Datasets:} 
Table~\ref{DataDescription} summarizes the datasets used in our experiments.
(1) Hot Roll of Steel (HRS): The hot rolling process in steel manufacturing requires monitoring conveyor rollers to prevent billet deformation, surface defects, downtime, and increased costs. This dataset, from an actual hot rolling steel industry, includes data from 294 rollers recorded via motor current signals and converted to digital data. It encompasses five fault categories: roller swing, roller stuck, overcurrent, squeaking, and base deformation. Due to the random occurrence of these faults, the dataset is highly unbalanced and contains limited samples. 
(2) Tennessee-Eastman process (TEP): It is widely used to validate fault diagnosis methods. It includes 21 distinct fault categories with training and testing sets. In this study, 4320 normal samples and 800 samples for each fault are used. 
(3) CAR Learning to Act (CARLA): Yan et al. \cite{Yan} collected this dataset using Dosovitskiy's \cite{dosovitskiy} autonomous driving simulator. It features single-sensor and multi-sensor faults across three maps: rainy, cloudy, and sunny conditions. There are 9 categories of single-sensor faults and 4 categories of multi-sensor faults. Each map had a 30-minute simulation with sensor data sampled at 60Hz. 

\textbf{Testing Scenarios:} We conduct experiments on three datasets under two scenarios: class-incremental and variable working conditions. The proposed method supports online continuous learning, eliminating the need to divide data into training and testing sets. We first initialize the network using 1000 normal samples. As monitoring data arrives, labeled samples and reliable samples with predicted labels from previous tasks are used for model training while predicting the new data. 
For class-incremental scenarios, normal samples are randomly divided equally into each task, and each fault sample appears in each task in turn. For varying working conditions, only the CARLA dataset includes three conditions, while the HRS and TEP datasets each have one, as shown in Table~\ref{DataDescription}. We gradually introduce noise into the HRS and TEP samples to simulate different working conditions. In the CARLA dataset, monitoring data evolve randomly from condition 1 to condition 2, and then to condition 3.

\textbf{Competing Methods:} We selected five methods, including state-of-the-art FD and four advanced OCL approaches, as benchmark algorithms for performance comparison: 1) Baseline: A basic model trained using experience replay (ER)~\cite{ChaudhryA}, achieving online continuous learning by replaying examples from previous tasks. 2) Adversarial Shapley Value Experience Replay (ASER)~\cite{Shimd}: Maintains learning stability and optimizes new class boundaries in the online class-incremental setting. 3) Camel~\cite{Li}: An advanced OCL method that accelerates model training and improves data efficiency through coreset selection. 4) Averaged Gradient Episodic Memory (AGEM)~\cite{Chaudhry}: Evaluates OCL efficiency in terms of sample complexity, computational cost, and memory usage. 5) MPOS-RVFL~\cite{HanPY}: An advanced ML-based FD method focused on real-time fault diagnosis with imbalanced data.

\textbf{Implementations:} The transformer network structure is used for all DL-based benchmarks. The basic neural network consists of an encoder, decoder, transformer encoder as the feature extractor, and a fully connected layer as the predictor. The encoder's hidden layer dimensions are 500, 500, and 2000, respectively, and the decoder's dimensions are 2000, 500, and 500. 
Model training used a learning rate of 0.0001 with the SGD optimizer, a maximum of 200 epochs, and a batch size of 100. 

\subcompact
\subsection{Performance Comparison}
Given the imbalanced nature of the experimental datasets, accuracy is not reliable. Therefore, we used metrics specifically designed for imbalanced datasets: Recall, Precision, G-mean, and F1 score. Recall measures the proportion of actual positives correctly identified, while Precision measures the proportion of predicted positives that are correct. G-mean balances sensitivity and specificity, and the F1 score is the harmonic mean of Precision and Recall. We also compared model training time. Our results are averaged across all tasks after the final update, denoted as Avg-End-Rel for Recall, Avg-End-Pre for Precision, Avg-End-Gmean for G-mean, and Avg-End-F1 for F1 score.

\textbf{Class-incremental:} The comparison results for class-incremental methods are shown in the left part of Table \ref{Results}. Our method requires the least training time on all datasets and outperforms other DL-based methods. Although the ML-based method MPOS-RVFL requires less training time, its performance is significantly inferior to other DL methods. The ER method is competitive, outperforming our method by 2.3\% in the four metrics on the CARLA-S dataset. However, ER's training time is 4.56 times longer, at 222.44 seconds compared to our 48.40 seconds. These results highlight the effectiveness of \system in balancing high performance with reasonable training times across various datasets and conditions.

\begin{table*}[t]
\centering
\caption{Performance comparison of the \system and other approaches across three datasets in two scenarios.}
\resizebox{\linewidth}{!}{
\begin{tabular}{cc|c|ccccc|ccccc}
\hline\hline 
\multirow{2}{*}{Dataset} & \multirow{2}{*}{DL/ML-based} & \multirow{2}{*}{Methods} & \multicolumn{5}{c|}{Class-incremental} & \multicolumn{5}{c}{Variable working condition} \\
\cline{4-13}
& && Avg-End-Rel & Avg-End-Pre  & Avg-End-F1  &Avg-End-Gmean  &Training-Time (s) & Avg-End-Rel & Avg-End-Pre  & Avg-End-F1  &Avg-End-Gmean  &Training-Time (s)  \\
\hline

\multirow{6}{*}{HRS} &ML-based& MPOS-RVFL & 0.3533  & 0.4083  & 0.3760 & 0.3784  & 9.04 & 0.1269 & 0.1667 & 0.1441 &0.1454   &6.73\\  \cline{2-13}
&\multirow{5}{*}{DL-based} & ASER & 0.3169 & 0.3750 & 0.2986 & 0.3397 & 355.35 &\underline{0.1937} & \textbf{0.5990} &\underline{0.2320}   &\underline{0.3373}   &225.72 \\ 
&& Camel & 0.5795  & 0.4926  & 0.5082  & 0.5327  & \underline{114.64} & 0.1520 & 0.5025 & 0.2185  &0.2762   &\underline{91.29}\\ 
&& AGEM & \underline{0.6032}  & \underline{0.5211}  & \underline{0.5382}  & \underline{0.5592}  & 176.79 &0.1468   &\underline{0.5597}   &0.2142    &0.2866  &117.98\\ 
&& ER & 0.5331 & 0.4594  & 0.4659 & 0.4935  & 144.48  & 0.1660& 0.5468   &0.1803 &0.2937  &93.75  \\ 
 && \system & \textbf{0.6831}  & \textbf{0.5663}  & \textbf{0.5881}  & \textbf{0.6187}  & \textbf{56.17} & \textbf{0.5914} & 0.5215 & \textbf{0.5365}    &\textbf{0.5552}   &\textbf{59.42}\\
\hline \hline 

\multirow{6}{*}{TEP} &ML-based& MPOS-RVFL & 0.0992 & 0.1678 & 0.1151 & 0.1228 & 14.33 & 0.0093 & 0.0455 & 0.0154 & 0.0205 & 4.27 \\ \cline{2-13}
&\multirow{5}{*}{DL-based}  & ASER & 0.1524 & 0.0737 & 0.0857 & 0.1042 & 348.72 & 0.3259 & 0.3258 & 0.2985  &0.3258  &350.97 \\ 
&& Camel & 0.1619 & 0.1173 & 0.1263 & 0.1359 & \underline{116.29} & 0.2364 & 0.2773 & 0.2289 & 0.2559 & \textbf{105.18} \\ 
&& AGEM & 0.1534 & 0.0993 & 0.1029 & 0.1208 & 139.47 & 0.2975 & 0.2643 & 0.2453 & 0.2803 & 131.21 \\ 
&& ER & \underline{0.1721} & \underline{0.1471} & \underline{0.1491} & \underline{0.1567} & 121.97 & \underline{0.3351} & \underline{0.3509} & \underline{0.3219} & \underline{0.3428} & 127.91 \\ 
&& \system & \textbf{0.1950} & \textbf{0.1606} & \textbf{0.1572} &\textbf{ 0.1742} & \textbf{77.10} & \textbf{0.3450} & \textbf{0.3569} & \textbf{0.3358} & \textbf{0.3508} & \underline{125.44} \\
\hline
\hline

\multirow{6}{*}{CARLS-S} &ML-based & MPOS-RVFL & 0.2694  & 0.2929  & 0.2787 & 0.2804  & 20.57 & 0.3755 & 0.4183 & 0.3942   &0.3955 & 29.80\\  \cline{2-13}
&\multirow{5}{*}{DL-based} & ASER & 0.2749 & 0.2642 & 0.2179 & 0.2579 & 563.18 & 0.5155 & 0.4378 & 0.4366 & 0.4718 & 1181.92 \\ 
&& Camel & 0.3896  & 0.3657  & 0.3274  & 0.3711  & \underline{156.65}   & 0.5733 & 0.5025 & 0.4925  &0.5319   &\underline{357.42}\\ 
&& AGEM & 0.4706  & 0.4501  & 0.4178  & 0.4569  & 271.62 & \textbf{0.6249} & 0.5031 & 0.4997  &0.5543   & 560.76\\ 
&& ER & \underline{0.5295} & \textbf{0.5916}  &\textbf{ 0.5077} & \textbf{0.5587}  & 222.44  &0.5907 & \textbf{0.5482}  &\textbf{0.5550}  &\textbf{0.5682}    &456.87\\ 
&& \system & \textbf{0.5519} & \underline{0.5160}  & \underline{0.4972}  & \underline{0.5306} &\textbf{48.04} & \underline{0.5982} & \underline{0.5115} & \underline{0.5117}   &\underline{0.5521}   &\textbf{95.99}\\
\hline
\hline

\multirow{6}{*}{CARLS-M} &ML-based & MPOS-RVFL & 0.4410 & 0.4567 & 0.4484 & 0.4486 & 29.89 & 0.4608 & 0.4833 & 0.4715 & 0.4718 & 28.92 \\ \cline{2-13}
&\multirow{5}{*}{DL-based} & ASER & 0.3701 & 0.4042 & 0.2625 & 0.3723 & 525.43 & 0.6985 & 0.6428 & 0.6488 & 0.6679 & 621.28 \\ 
&& Camel & 0.5610 & 0.5303 & 0.5021 & 0.5388 & \underline{148.50} & 0.5775 & 0.4836 & 0.4720 & 0.5221 & \underline{230.10} \\ 
&& AGEM & 0.5503 & 0.5183 & 0.5166 & 0.5326 & 262.09 & 0.6062 & 0.4068 & 0.3983 & 0.4784 & 355.06 \\  
&& ER & \textbf{0.6269} & \underline{0.6403} & \underline{0.6172} & \underline{0.6332} & 229.13 & \underline{0.7856} & \underline{0.6886} & \underline{0.7109} & \underline{0.7330} & 281.76 \\ 
&& \system & \underline{0.6108} & \textbf{0.7133} & \textbf{0.6400} & \textbf{0.6568} & \textbf{45.57} & \textbf{0.8223} & \textbf{0.7089} & \textbf{0.7314} & \textbf{0.7617} & \textbf{37.35} \\   
\hline
\hline
\end{tabular}
}
\label{Results}
\end{table*}

\begin{table*}[t]
\centering
\caption{Performance of \system compared to its variants without each component across three datasets.}
\resizebox{\linewidth}{!}{
\begin{tabular}{cc|ccccc|ccccc}
\hline
\multirow{2}{*}{Dataset} & \multirow{2}{*}{Methods} & \multicolumn{5}{c|}{Class-incremental} & \multicolumn{5}{c}{Variable working condition} \\
\cline{3-12}
& & Avg-End-Rel & Avg-End-Pre  & Avg-End-F1  &Avg-End-Gmean  &Training-Time (s) & Avg-End-Rel & Avg-End-Pre  & Avg-End-F1  &Avg-End-Gmean  &Training-Time (s)  \\
\hline

\multirow{6}{*}{HRS}  & \system w/o RSC & 0.6034 & \underline{0.5619} &\underline{0.5725} & 0.5815 & 362.29 &\textbf{0.7196} & \underline{0.6808} &\textbf{0.6862}   &\textbf{0.6998}   &307.35 \\ 
& \system w/o GBT & \underline{0.6437} & 0.5464 & 0.5699 & \underline{0.5919} & 218.17 &\underline{0.6750} & \textbf{0.7005} &\underline{0.6858}   &\underline{0.6875}   &174.93\\ 
& \system w/o CUPL & 0.5897 & 0.5003 & 0.5127 & 0.5405 & \textbf{56.00} &0.4109 & 0.4311 &0.4019   &0.4198   &\textbf{18.08}\\ 
& \system  &\textbf{0.6831}  &\textbf{0.5663}  &\textbf{0.5881}  &\textbf{0.6187}  &\underline{56.17} &0.5914 & 0.5215 &0.5365    &0.5552   &\underline{59.42}\\
 \hline 

\multirow{6}{*}{TEP}  & \system w/o RSC & \textbf{0.2174} &\underline{0.1642} &\underline{0.1743} &\textbf{0.1880} & 168.05 &\underline{0.3203} &\underline{0.3469} &\underline{0.3119}   &\underline{0.3332}   &159.60 \\ 
& \system w/o GBT & 0.2053 & \textbf{0.1889} &\textbf{0.1793} & \underline{0.1857} & 140.25 &0.2036 & 0.2105 &0.1919   &0.2070   &115.17\\ 
& \system w/o CUPL & 0.1055 & 0.0824 & 0.0803 & 0.0915 &\underline{87.94} &0.2343 & 0.2615 &0.2221   &0.2474   &\textbf{65.86}\\ 
& \system   &\underline{0.1950} &0.1606 &0.1572 &0.1742 &\textbf{77.10} &\textbf{0.3450}&\textbf{0.3569}&\textbf{0.3358} &\textbf{0.3508} &\underline{125.44}\\
 \hline 

\multirow{6}{*}{CARLS-S}& \system w/o RSC & 0.5133 & \underline{0.5169} & 0.4769 & 0.5135 & 334.64 &\underline{0.6878} &\textbf{0.5972} &\textbf{0.6133}   &\underline{0.6390}   &986.64 \\ 
& \system w/o GBT &\textbf{0.5922} & \textbf{0.5518} &\textbf{0.5410} &\textbf{0.5702} & 222.49 &\textbf{0.7340} &\underline{0.5878} &\underline{0.6098}   &\textbf{0.6526}   &689.73\\ 
& \system w/o CUPL & 0.3907 & 0.4732 & 0.3684 & 0.4265 &\underline{61.42} &0.3620 & 0.3302 &0.3221   &0.3450   &\textbf{34.19}\\ 
& \system  &\underline{0.5519} & 0.5160  & \underline{0.4972}  & \underline{0.5306} &\textbf{48.04} &0.5982 & 0.5115 &0.5117   &0.5521   &\underline{95.99}\\
 \hline 

\multirow{6}{*}{CARLS-M} & \system w/o RSC &\underline{0.6222} & 0.6360 & 0.6121 & 0.6262 & 673.35 &0.7905 &\underline{0.6881} &\textbf{0.7134}   &\underline{0.7363}   &300.65 \\ 
& \system w/o GBT &\textbf{0.7389} &\textbf{0.7149} & \textbf{0.7176} &\textbf{0.7255} & 190.73 &\textbf{0.8236} & 0.6199 &0.6379   &0.7044   &99.53\\ 
& \system w/o CUPL & 0.5104 & 0.4865 & 0.4446 & 0.4939 &\underline{55.39} &0.7764 & 0.6644 &0.6906   &0.7168   &\underline{44.34}\\ 
& \system  & 0.6108 &\underline{0.7133} &\underline{0.6400} &\underline{0.6568} & \textbf{45.57} & \underline{0.8223} &\textbf{0.7089} &\textbf{0.7314} &\textbf{0.7617} & \textbf{37.35} \\   
 \hline 
\end{tabular}
}
\label{Results}
\end{table*}

\textbf{Variable working conditions:} The right part of Table \ref{Results} shows the performance comparison under variable working conditions across three datasets. \system consistently outperforms other approaches, achieving the highest scores in most performance metrics. While the ER method outperforms our method on the CARLS-S dataset, it requires much more training time (ER: 456.87 seconds vs. \system: 95.99 seconds). The CAMEL method maintains the second shortest training time across all datasets, leading on the TEP dataset with 105.18 seconds, 20.26 seconds shorter than \system (125.44 seconds). However, CAMEL's overall performance is 9.7\% lower than \system in other metrics. These results highlight the effectiveness of \system in balancing high performance with reasonable training times across various datasets and conditions.


\subcompact
\subsection{Ablation Study}
To evaluate the contribution of each component in \system, we separately removed the RSC, GBT, and CUFL components and conducted experiments on all datasets. We compared the performance before and after removing each component under class-incremental and variable working conditions. From Table 4, the following three conclusions can be drawn.

First, the RSC module effectively reduces model training time. When comparing \textit{SRTFD} with \textit{SRTFD w/o RSC}, the training duration of \textit{SRTFD} is significantly shortened across all datasets while maintaining comparable performance across four metrics. For instance, in the HRS dataset within a class-incremental setting, the training time is reduced by 80.66\%, and the Avg-End-rel improves by 13.22\%.

Second, the GBT module contributes to reducing model training time and improving FD performance. The GBT module, which considers balanced coreset selection, is integrated with the RSC module responsible for coreset selection. As a result, \textit{SRTFD w/o GBT} performs better than \textit{SRTFD} due to training on more data, but requires more training time. For example, in the TEP dataset within a class-incremental setting, \textit{SRTFD w/o GBT} increases Avg-End-Rel by 1.03\% but decreases training time by 45.26\%. However, in the HRS and TEP datasets, and in some cases of the CARLS data, \textit{SRTFD w/o GBT} performs worse than \textit{SRTFD w/o RSC}. For instance, Avg-End-Pre of \textit{SRTFD w/o GBT} is 0.5878, while \textit{SRTFD w/o RSC} is 0.5972. This shows the effectiveness of the GBT module in handling imbalanced data.

Lastly, the CUPL module significantly improves FD performance. While the training time for \textit{RSTFD w/o CUPL} may be shorter in certain cases, it is evident that its performance is markedly inferior to that of \textit{RSTFD}. A comparison between \textit{RSTFD} and \textit{RSTFD w/o CUPL} reveals that the extensive use of pseudo-labeled samples for model training substantially improves FD performance. For instance, the Avg-End-Rel of \textit{RSTFD} increased by 9.34\% in the HRS dataset within a class-incremental setting, while the training times are similar (56.17 vs. 56.00 seconds).

\subcompact
\subsection{Analysis and Discussion}
\subsubsection{Effects of coreset ratio and cluster number \(uc\)}
The impact of these two parameters is illustrated in Figure \ref{RCS1} and Figure \ref{RCS} in the appendix. Panels (a-d) in Figure \ref{RCS} show the performance across all datasets in the class-incremental scenario for coreset ratios of [0.5, 0.6, 0.7, 0.8, 0.9]. From this figure, it can be observed that as the coreset ratio increases, the number of samples in the coreset grows, leading to longer training times for the model. Additionally, when the coreset ratio is 0.6, the performance of the HRS, CARLS-S, and CARLS-M models is relatively better, and the training time is shorter. 
Additionally, Figures \ref{RCS} (e-h) demonstrate the model performance with different numbers of clusters, uc, set to [3, 6, 9, 12, 15]. From the figures, it can be observed that although the training performance of the models is relatively stable across different uc values, the choice of uc is related to the number of categories in the dataset. For instance, the HRS dataset has 5 categories, while the TEP dataset has 22 categories. Therefore, on the HRS dataset, a uc value of 3 achieves a trade-off in performance across the four metrics and training time, whereas on the TEP dataset, a uc value of 12 achieves a balance.
\begin{figure}[t]
    \centering
    \begin{subfigure}[b]{0.23\textwidth}
        \centering
        \includegraphics[width=\textwidth]{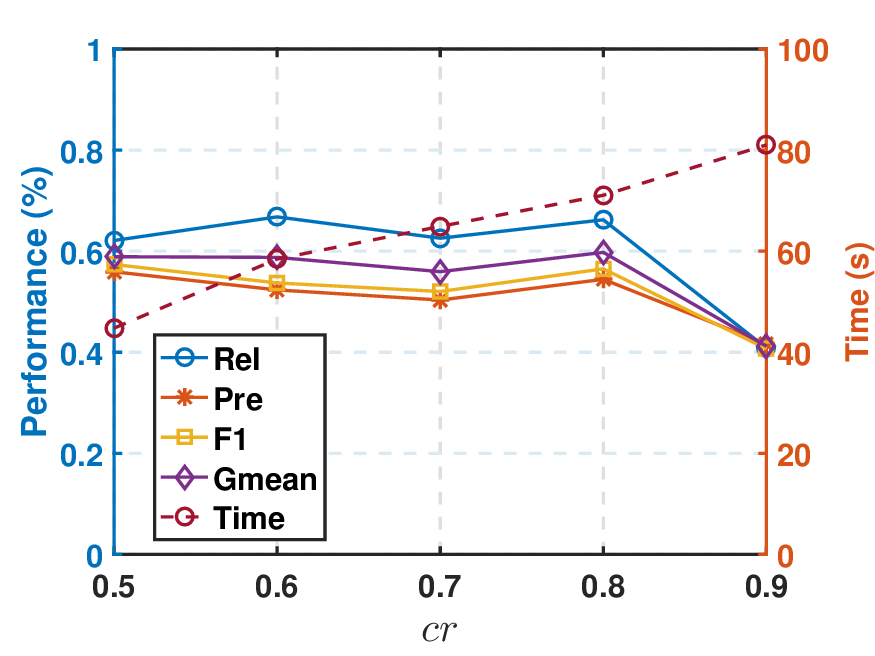}
        \caption{Effect of varying coreset ratio \(cr\).}
    \end{subfigure}
    \begin{subfigure}[b]{0.23\textwidth}
        \centering
        \includegraphics[width=\textwidth]{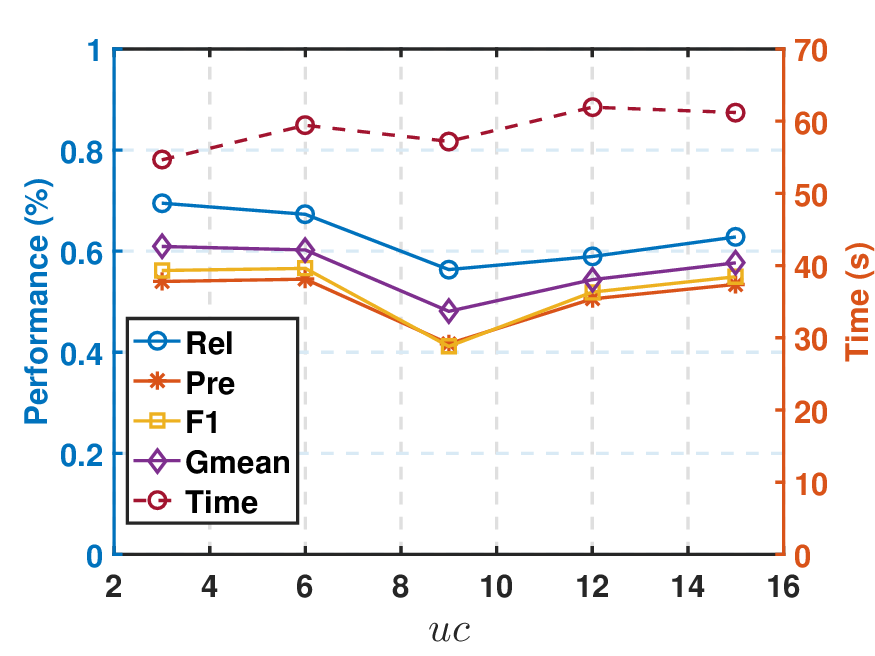}
        \caption{Effect of varying cluster number \(uc\).}
    \end{subfigure}

    \caption{Performance of different coreset ratios \(cr\) and cluster numbers \(uc\) on HRS datasets within class-incremental.}
    \label{RCS1}
\end{figure}

\begin{figure}[t]
    \centering

    \begin{subfigure}[b]{0.23\textwidth}
        \centering
        \includegraphics[width=\textwidth]{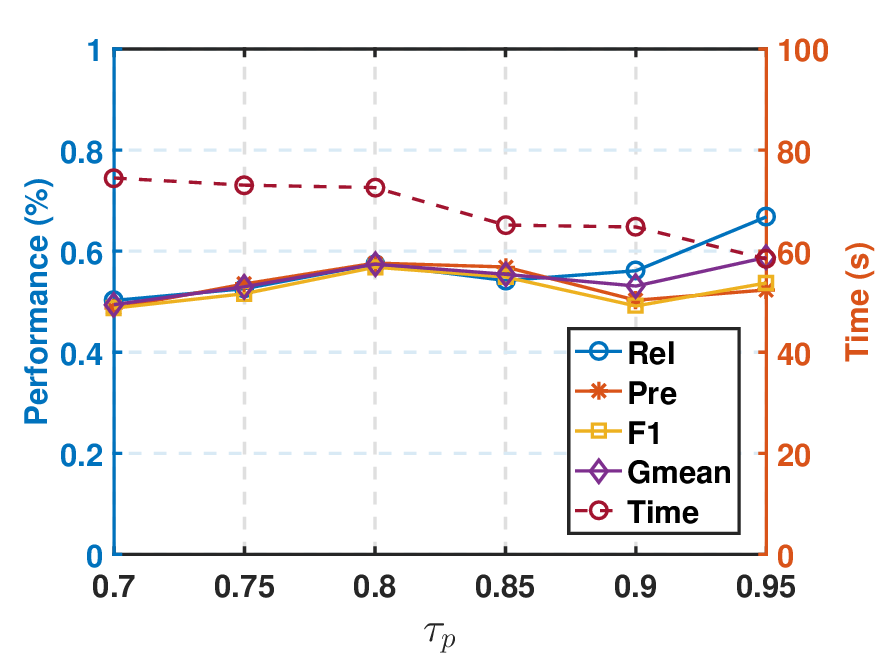}
        \caption{Effect of varying positive label threshold \(\tau_p\).}
    \end{subfigure}
    \begin{subfigure}[b]{0.23\textwidth}
        \centering
        \includegraphics[width=\textwidth]{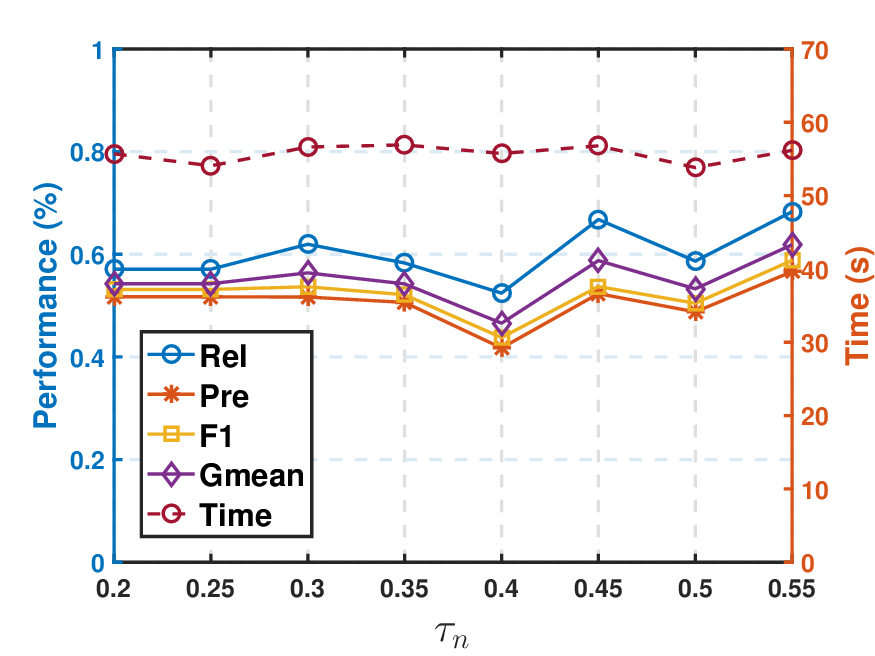}
        \caption{Effect of varying negative label threshold \(\tau_n\).}
    \end{subfigure}

        \begin{subfigure}[b]{0.23\textwidth}
        \centering
        \includegraphics[width=\textwidth]{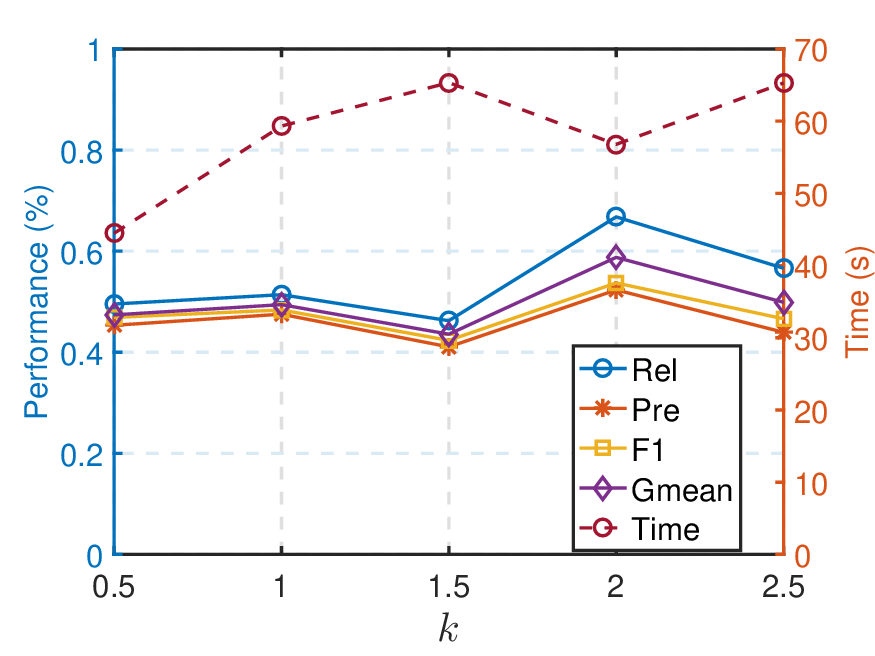}
        \caption{Effect of varying model prediction threshold \(k\).}
    \end{subfigure}
        \begin{subfigure}[b]{0.23\textwidth}
        \centering
        \includegraphics[width=\textwidth]{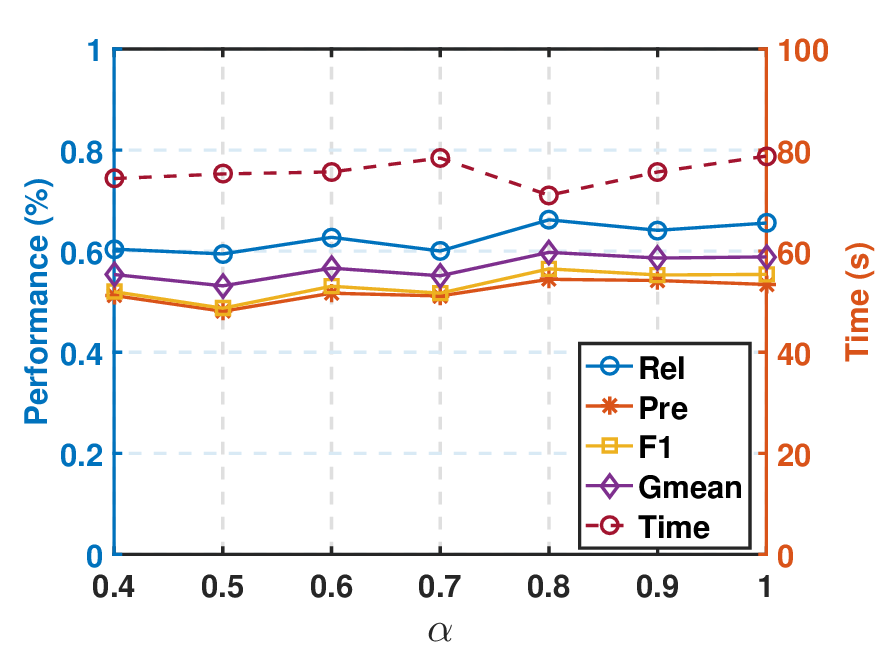}
        \caption{Effects of varying pseudo-labeled sample weights \(\alpha\).}
    \end{subfigure}

    \caption{Performance of different thresholds and weights on HRS datasets within class-incremental setting.}
    \label{CUPL1}
\end{figure}

\subsubsection{Effects of parameters for pseudo-label samples selection}
For the selection of pseudo-label samples, there are three thresholds, positive label threshold \(\tau_p\), negative label threshold \(\tau_n\), and uncertainty threshold of model prediction \(k\). Figure \ref{CUPL1} (a, b, c) and Figure \ref{CUPL} in the appendix demonstrate the performance of these three thresholds on different datasets. Regarding the positive label threshold, as shown in Figures \ref{CUPL} (a-d), higher values lead to better model performance due to using pseudo-labels with higher confidence. For the negative label threshold, as shown in Figure \ref{CUPL} (e-h), a value of 0.45 results in the best performance across all datasets. When the threshold exceeds 0.45, the model performance tends to decline, as illustrated in Figure \ref{CUPL} (f) for the TEP dataset. The impact of the uncertainty threshold on model performance is shown in Figures \ref{CUPL} (i-l). When the value is set to 2, the model performs better across all datasets. 

\subsubsection{Effects of pseudo-labeled sample weights \(\alpha\)}
Figure \ref{CUPL1} (d) and Figure \ref{alph} in the appendix show the performance of different pseudo-labeled sample weights \(\alpha\) on all datasets within class-incremental setting. From Figure \ref{alph}, it can be observed that changes in the value of alpha do not significantly impact the model training time. However, there are minor effects on the model's Recall, Precision, F1, and Gmean metrics. The optimal performance is achieved with an \(alpha\) value of 0.7 on the HRS, TEP, and CARLS-M datasets, while an \(alpha\) value of 0.6 performs best on the CARLS-S dataset.

\compact
\section{Related Work}
\paragraph{Fault Diagnosis}
Current state-of-the-art fault diagnosis (FD) techniques, summarized in Table~\ref{Existing_FD}, fail to meet real-world demands. Modern FD methods must adapt to new fault classes and conditions, handle scalable data, provide real-time responses, and function with minimal prior information, such as limited labeled samples. Despite advances, existing FD methods face significant challenges. For instance, OSELM~\cite{Haow}, D-CART~\cite{Denghx}, and TCNN~\cite{Xu} struggle with data efficiency and scalability. OSSBLS~\cite{Puxk} and ODDFD~\cite{Linjh} manage limited labeled data but fail to address data imbalance effectively. AMPNet~\cite{Fedor} handles large-scale data but lacks adaptability to new fault types and changing environments. Additionally, 1D-CNN~\cite{Elsisi} faces significant challenges with scalability. 
OLFA~\cite{Yan} manages limited labeled data but does not effectively address real-time response and high accuracy. Real-time response capabilities are crucial, yet only TCNN~\cite{Xu} shows low training costs conducive to quick responses, unlike AMPNet~\cite{Fedor} and OLFA~\cite{Yan}. Effective diagnosis with minimal labeled samples remains a challenge for many methods. High performance and low training costs are essential but not prioritized by several methods, making them impractical in real-world applications. Thus, there is a critical need for advanced, scalable, and precise FD techniques that address these challenges effectively.
In parallel, recent advancements in computer vision have focused on industrial anomaly detection (IAD)~\cite{Alex, Aimira, WangCJ, FangZ}. However, existing IAD approaches are unsuitable for FD due to fundamental differences. FD targets specific functional problems using sensor data~\cite{Fedor}, while IAD identifies statistical features and pattern changes indicating abnormalities using image data~\cite{Bae, WangY}. Thus, the clear distinction of use cases drives also a clear separation of concerns.


\vspace{-4pt}
\paragraph{Online Continual Learning} 
OCL aims to develop models that can continuously learn from a stream of data 
without forgetting previously acquired knowledge~\cite{Aljundi}. Key methods include replay-based techniques like experience replay (ER)~\cite{ChaudhryA} and adversarial shapley value experience Replay (ASER)~\cite{Shimd}, which store and replay subsets of past data. Regularization approaches such as gradient
episodic memory (GEM)~\cite{Lopez} and averaged GEM (AGEM)~\cite{Chaudhry} add constraints to protect important weights. Parameter isolation methods like progressive neural networks~\cite{Rusu} and PackNet~\cite{Mallya}, allocate different model parameters to different tasks. Coreset selection methods like GoodCore~\cite{Chai} and Camel~\cite{Li}, further address scalability and memory efficiency issues of OCL by selecting representative data subsets. While OCL provides a robust framework, it requires significant adaptation to effectively meet the unique demands of real-time fault diagnosis in modern industrial environments.
\compact
\section{Conclusion}
This paper addresses high training costs, limited labeled samples, and imbalance issues in scalable real-time fault diagnosis using online continual learning with coreset selection. The proposed \system framework comprises three key components: \textit{RCS}, \textit{GBT}, and \textit{CUPL}. These innovations enhance fault diagnosis by improving model performance and efficiency in industrial settings. Extensive validation on real-world and simulated datasets shows significant improvements in four metrics while reducing training time. Our approach tackles data redundancy and imbalance, providing cost-efficient and robust solutions for industrial fault diagnosis. Future work will optimize the \system framework and explore its applicability to other industrial processes and environments.
\bibliographystyle{ACM-Reference-Format}
\bibliography{mybib}

\newpage
\appendix
\section{Reproducibility}
\textbf{1. Environment Requirements}

Ensure you have all the necessary Python packages by installing them from the provided `requirements.txt` file.

\textbf{2. Data Sources}

- HRS Dataset: This dataset is private and specific to the requirements of cooperative factories.

- TEP and CARLS Datasets: These two datasets are included in our publicly available code.

\textbf{3. Setting Up and Running SRTFD}

i. Install Required Packages

   Make sure you have Python installed. Then, navigate to the project directory and install the required packages using the following command:
   pip install -r requirements.txt
   
ii. Run SRTFD
   
   Execute the main script to start the SRTFD process:
   
python3 general\_main.py --data TEP --num\_tasks 22 --cl\_type nc --agent SRTFD --num\_runs 1 --N 1000

python3 general\_main.py --data TEP --num\_tasks 22 --cl\_type vc --agent SRTFD --num\_runs 1 --N 1000

python3 general\_main.py --data CARLS\_S --num\_tasks 10 --cl\_type nc --agent SRTFD --num\_runs 1 --N 1000

python3 general\_main.py --data CARLS\_S --num\_tasks 10 --cl\_type vc --agent SRTFD --num\_runs 1 --N 1000  

python3 general\_main.py --data CARLS\_M --num\_tasks 5 --cl\_type nc --agent SRTFD --num\_runs 1 --N 1000

python3 general\_main.py --data CARLS\_M --num\_tasks 5 --cl\_type vc --agent SRTFD --num\_runs 1 --N 1000

Additional Resources
For more detailed instructions and documentation, please refer to the project's test.bash file or the official documentation provided with the project.

\section{More Experimental Results}
\begin{figure*}[t]
    \centering
    \begin{subfigure}[b]{0.24\textwidth}
        \centering
        \includegraphics[width=\textwidth]{Figures/HRS/HRS_CoreRate.eps}
        \caption{HRS}
    \end{subfigure}
    \begin{subfigure}[b]{0.24\textwidth}
        \centering
        \includegraphics[width=\textwidth]{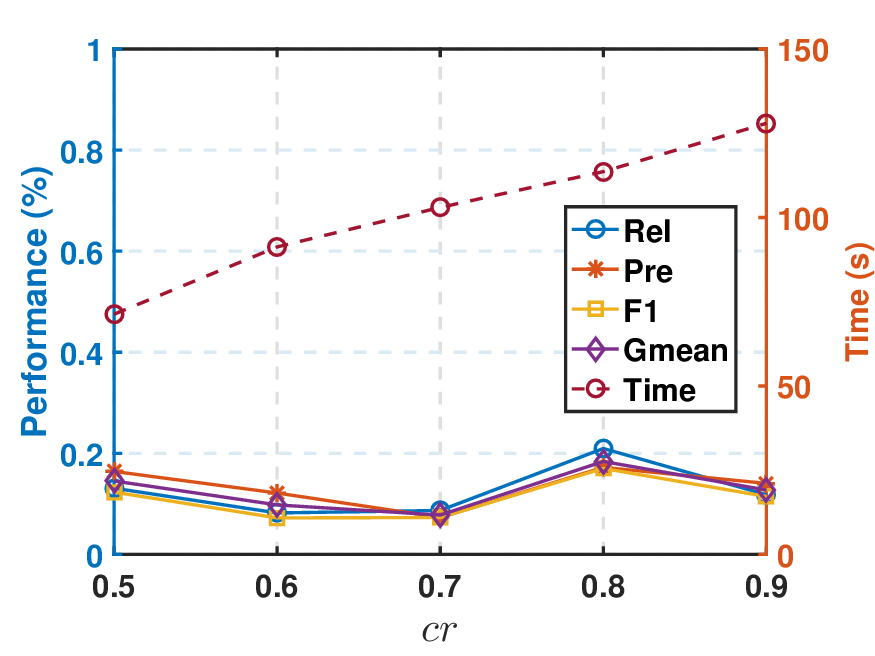}
        \caption{TEP}
    \end{subfigure}
    \begin{subfigure}[b]{0.24\textwidth}
        \centering
        \includegraphics[width=\textwidth]{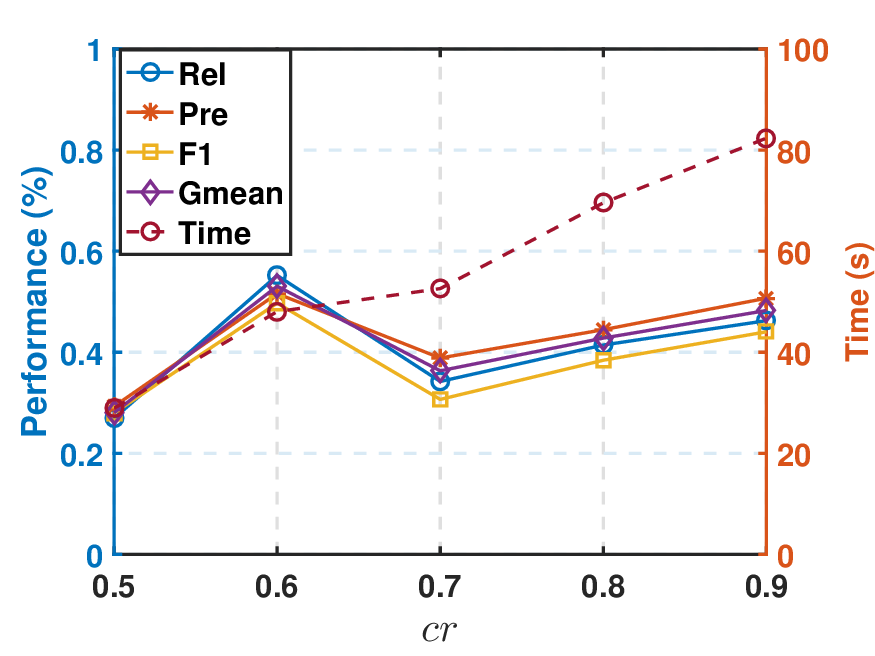}
        \caption{CARLS-S}
    \end{subfigure}
    \begin{subfigure}[b]{0.24\textwidth}
        \centering
        \includegraphics[width=\textwidth]{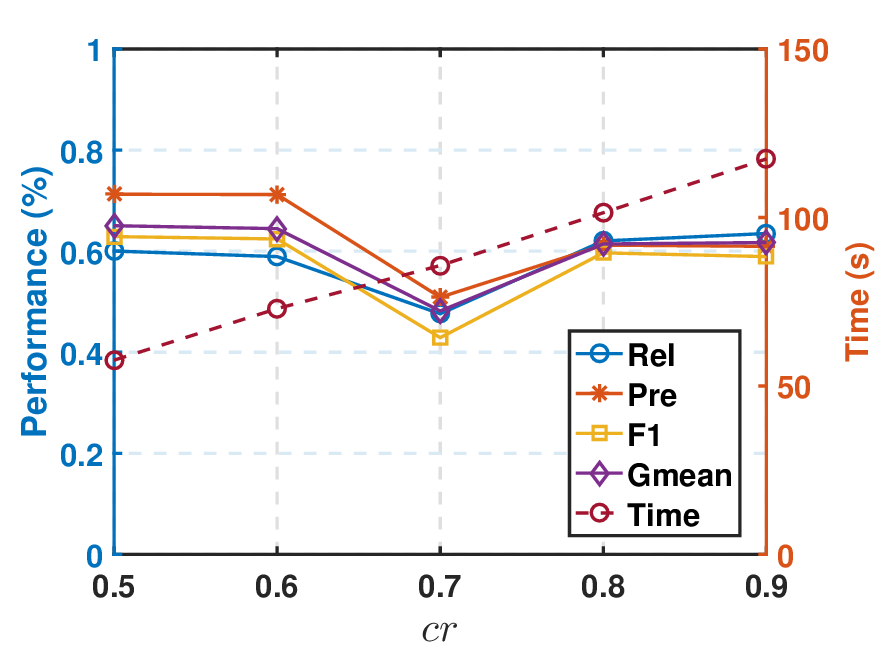}
        \caption{CARLS-M}
    \end{subfigure}
    \begin{subfigure}[b]{0.24\textwidth}
        \centering
        \includegraphics[width=\textwidth]{Figures/HRS/HRS_uc.eps}
        \caption{HRS}
    \end{subfigure}
    \begin{subfigure}[b]{0.24\textwidth}
        \centering
        \includegraphics[width=\textwidth]{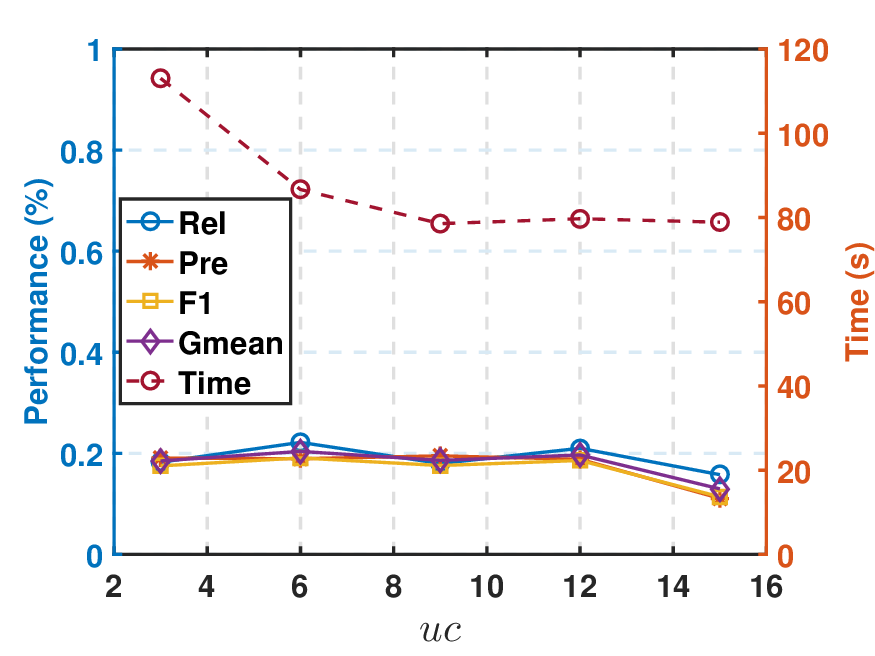}
        \caption{TEP}
    \end{subfigure}
    \begin{subfigure}[b]{0.24\textwidth}
        \centering
        \includegraphics[width=\textwidth]{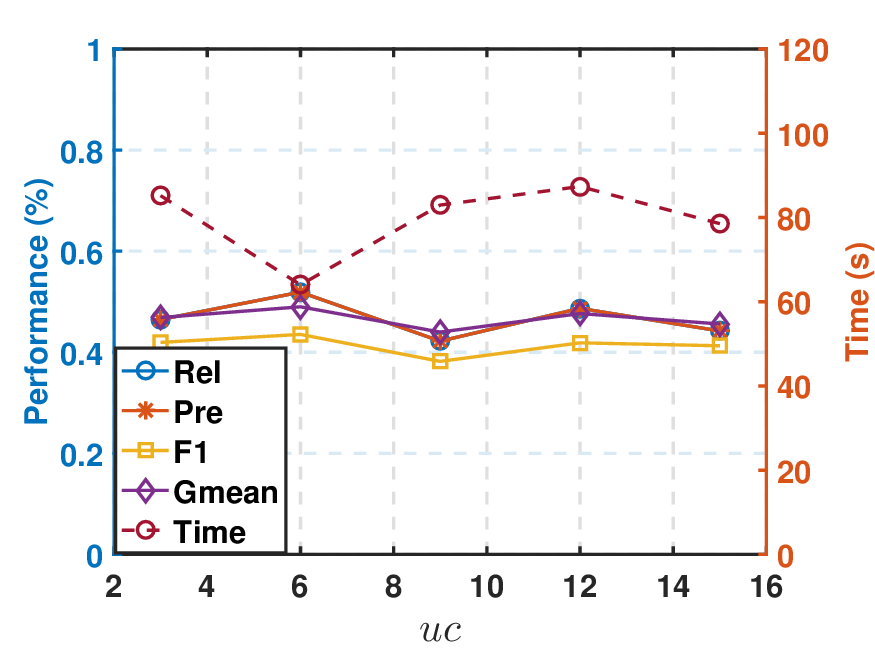}
        \caption{CARLS-S}
    \end{subfigure}
    \begin{subfigure}[b]{0.24\textwidth}
        \centering
        \includegraphics[width=\textwidth]{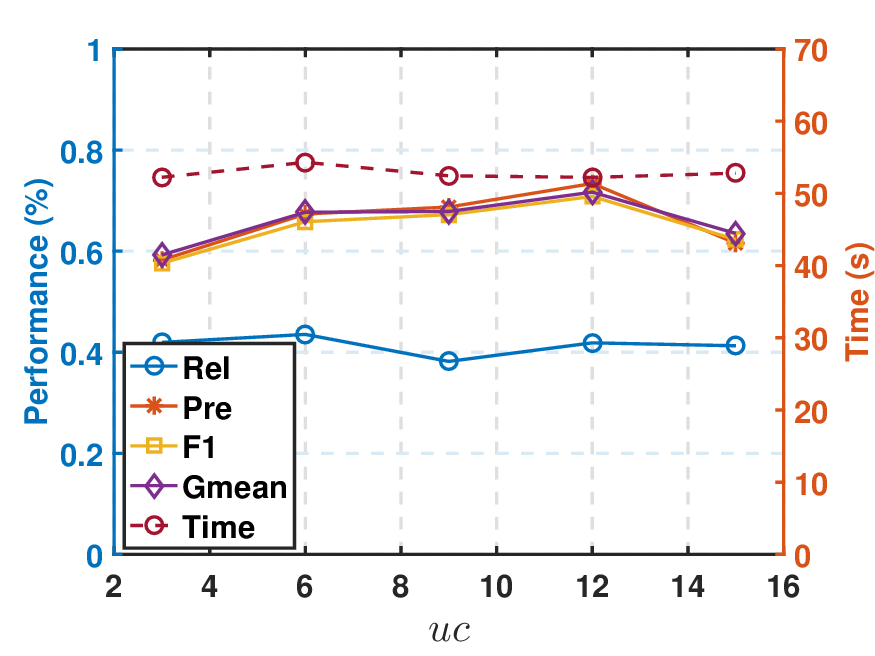}
        \caption{CARLS-M}
    \end{subfigure} 
    \caption{Performance of different coreset ratios and cluster numbers \(uc\) on all datasets within class-incremental.}
    \label{RCS}
\end{figure*}

\begin{figure*}[t]
    \centering

    \begin{subfigure}[b]{0.24\textwidth}
        \centering
        \includegraphics[width=\textwidth]{Figures/HRS/HRS_HighConfi.eps}
        \caption{HRS}
    \end{subfigure}
    \begin{subfigure}[b]{0.24\textwidth}
        \centering
        \includegraphics[width=\textwidth]{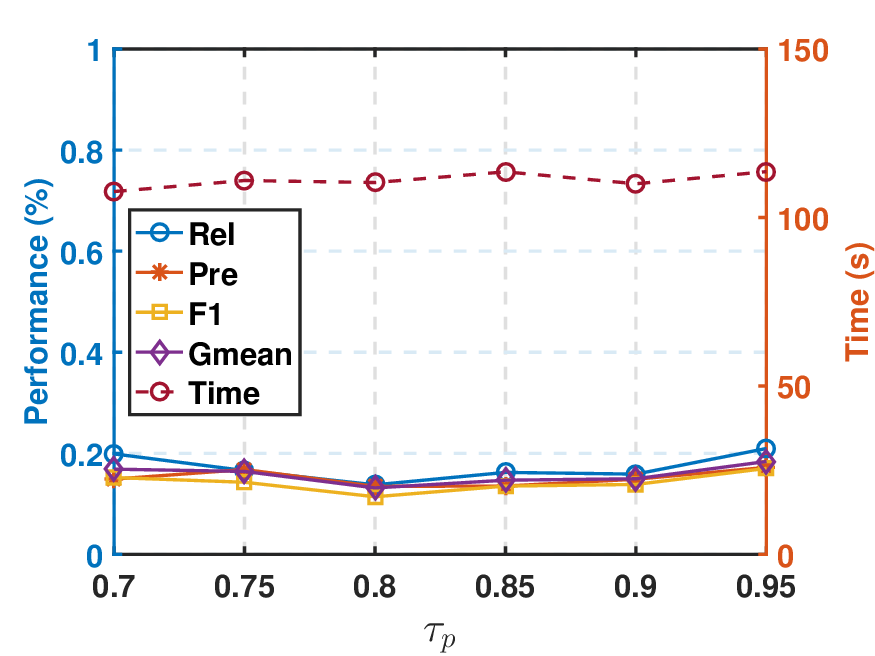}
        \caption{TEP}
    \end{subfigure}
    \begin{subfigure}[b]{0.24\textwidth}
        \centering
        \includegraphics[width=\textwidth]{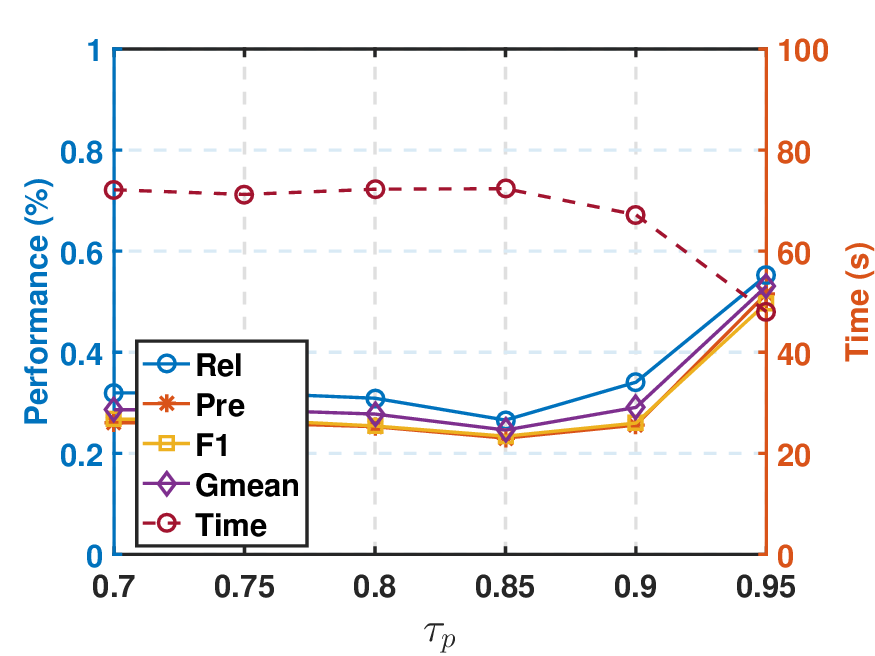}
        \caption{CARLS-S}
    \end{subfigure}
    \begin{subfigure}[b]{0.24\textwidth}
        \centering
        \includegraphics[width=\textwidth]{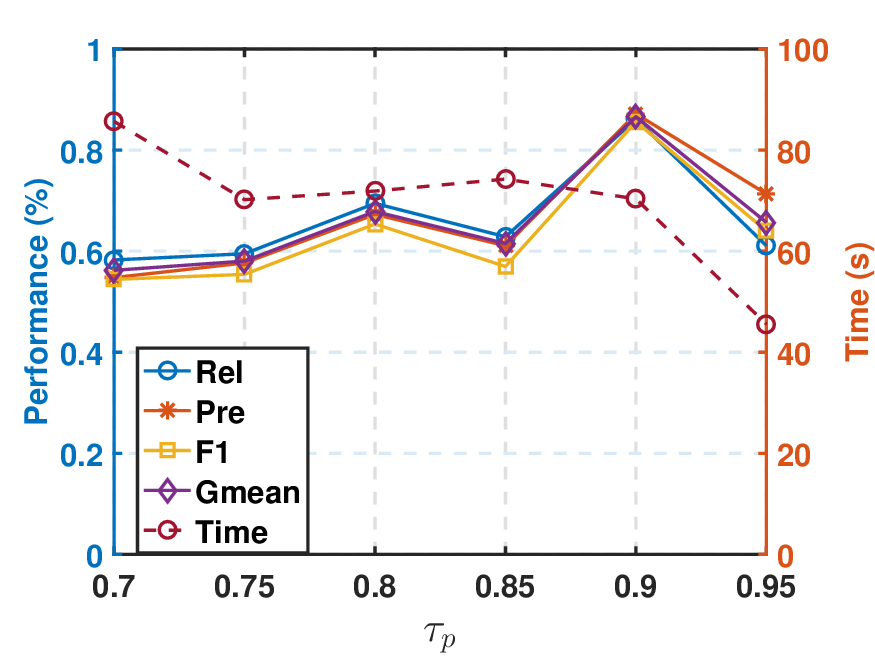}
        \caption{CARLS-M}
    \end{subfigure}
    
    \begin{subfigure}[b]{0.24\textwidth}
        \centering
        \includegraphics[width=\textwidth]{Figures/HRS/HRS_LowConfi.eps}
        \caption{HRS}
    \end{subfigure}
    \begin{subfigure}[b]{0.24\textwidth}
        \centering
        \includegraphics[width=\textwidth]{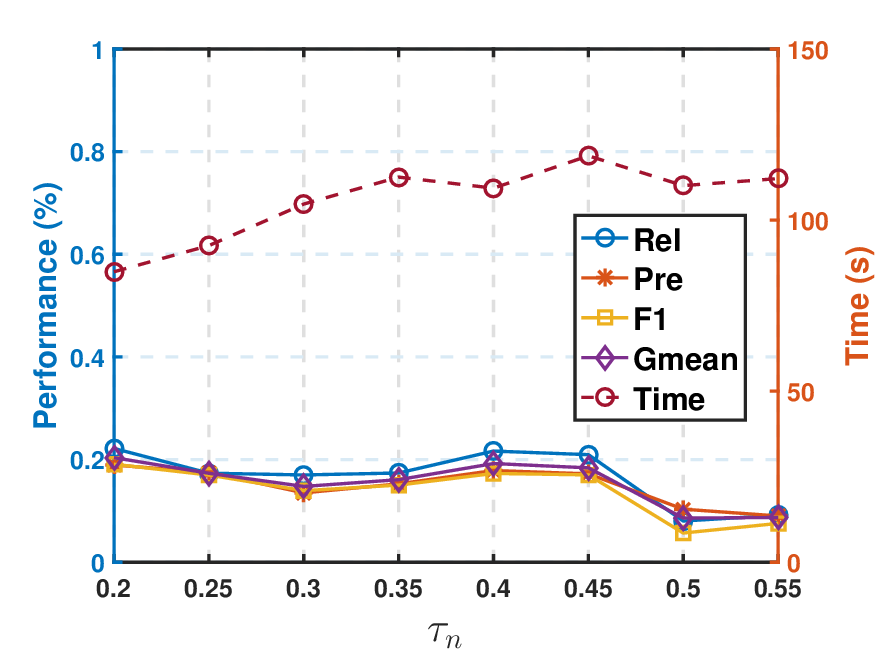}
        \caption{TEP}
    \end{subfigure}
    \begin{subfigure}[b]{0.24\textwidth}
        \centering
        \includegraphics[width=\textwidth]{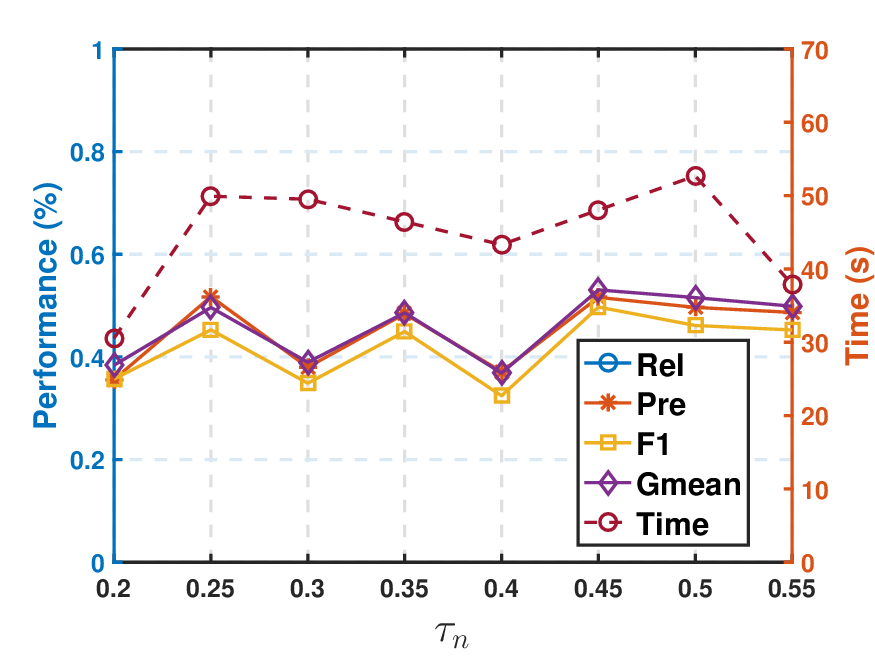}
        \caption{CARLS-S}
    \end{subfigure}
    \begin{subfigure}[b]{0.24\textwidth}
        \centering
        \includegraphics[width=\textwidth]{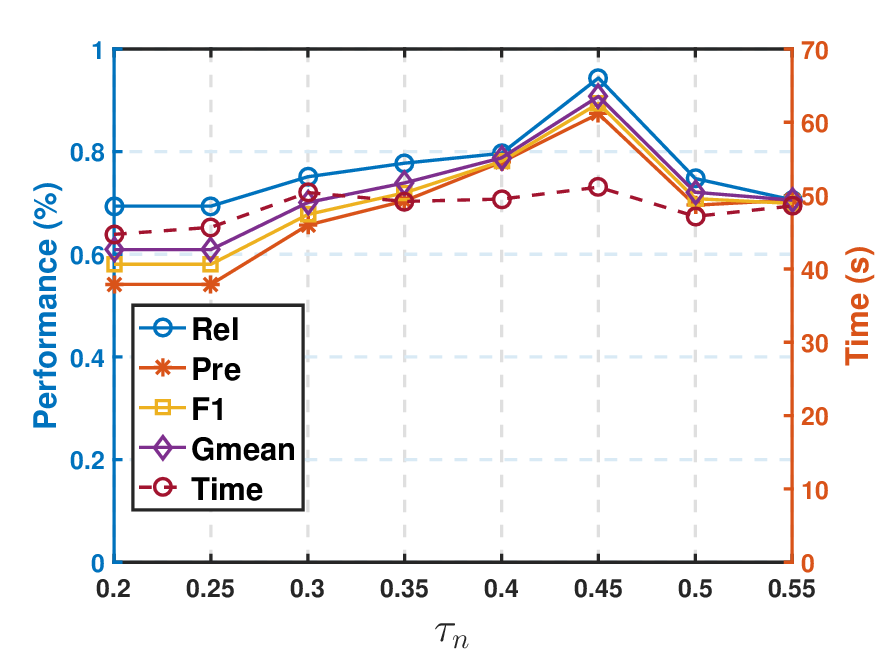}
        \caption{CARLS-M}
    \end{subfigure}
    
        \begin{subfigure}[b]{0.24\textwidth}
        \centering
        \includegraphics[width=\textwidth]{Figures/HRS/HRS_k.eps}
        \caption{HRS}
    \end{subfigure}
    \begin{subfigure}[b]{0.24\textwidth}
        \centering
        \includegraphics[width=\textwidth]{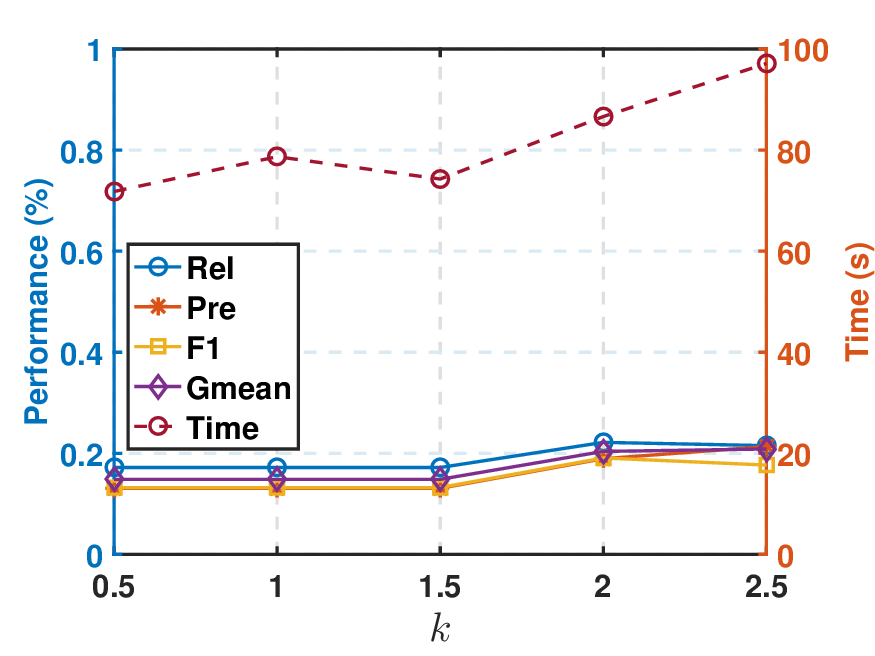}
        \caption{TEP}
    \end{subfigure}
    \begin{subfigure}[b]{0.24\textwidth}
        \centering
        \includegraphics[width=\textwidth]{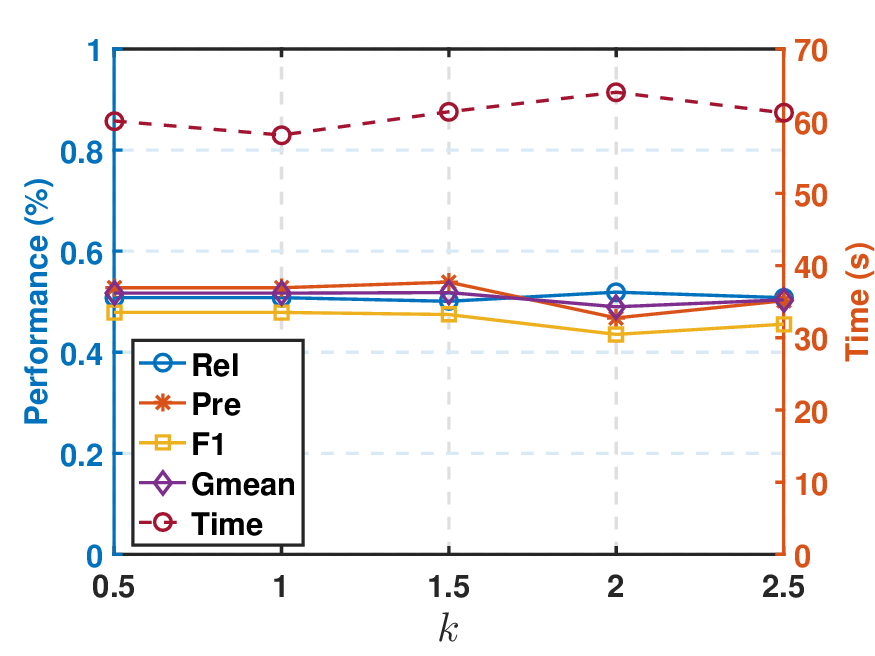}
        \caption{CARLS-S}
    \end{subfigure}
    \begin{subfigure}[b]{0.24\textwidth}
        \centering
        \includegraphics[width=\textwidth]{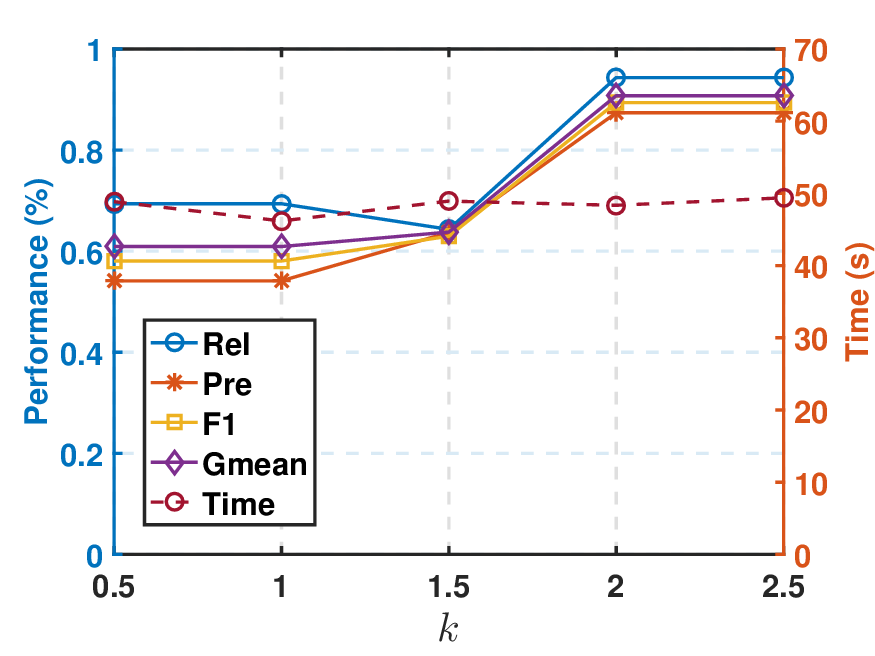}
        \caption{CARLS-M }
    \end{subfigure}
    
    \caption{Performance of different thresholds on all datasets within class-incremental setting.}
    \label{CUPL}
\end{figure*}

\begin{figure*}[t]
    \centering
    \begin{subfigure}[b]{0.24\textwidth}
        \centering
        \includegraphics[width=\textwidth]{Figures/HRS/HRS_Alpha.eps}
        \caption{HRS}
    \end{subfigure}
    \begin{subfigure}[b]{0.24\textwidth}
        \centering
        \includegraphics[width=\textwidth]{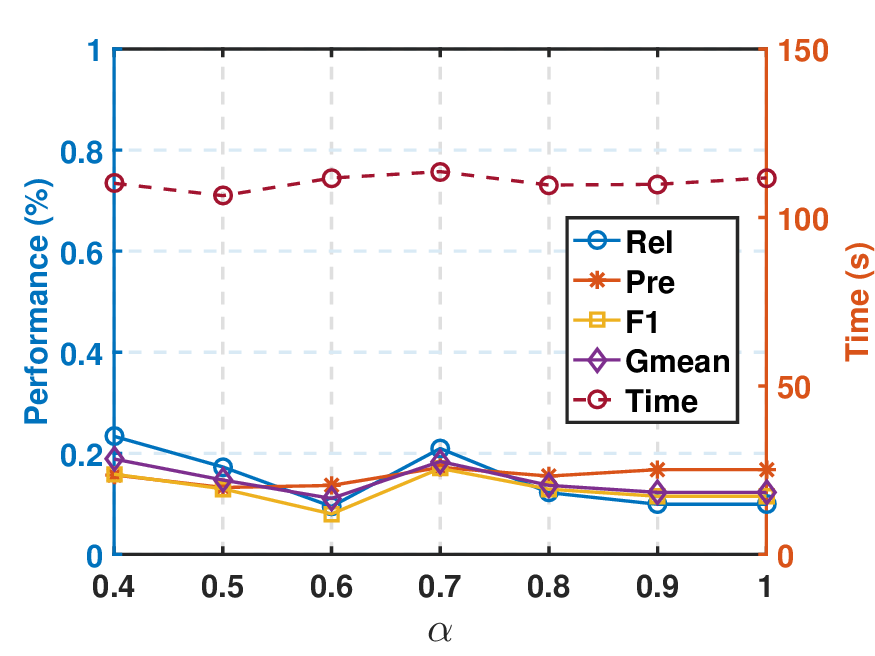}
        \caption{TEP}
    \end{subfigure}
    \begin{subfigure}[b]{0.24\textwidth}
        \centering
        \includegraphics[width=\textwidth]{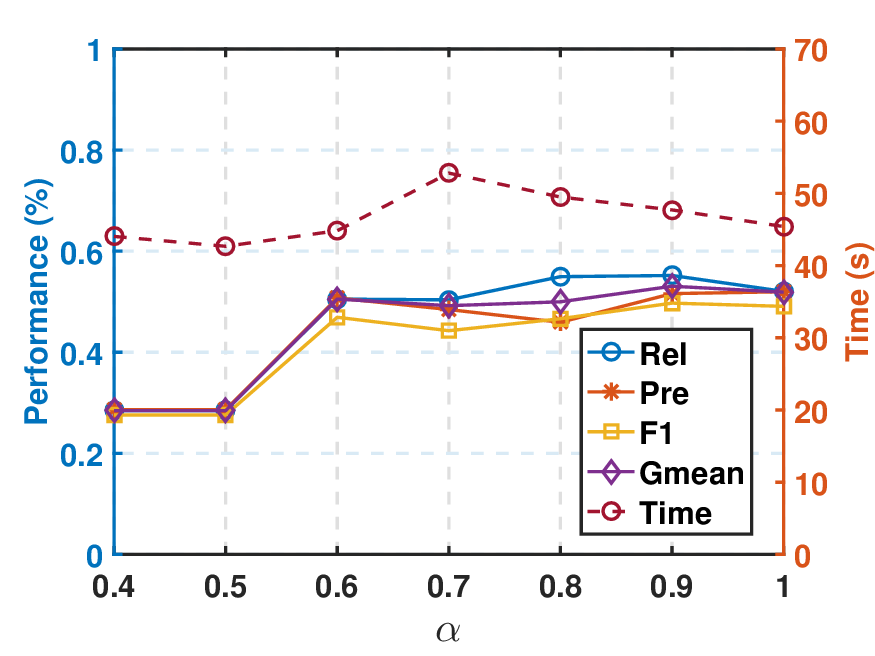}
        \caption{CARLS-S}
    \end{subfigure}
    \begin{subfigure}[b]{0.24\textwidth}
        \centering
        \includegraphics[width=\textwidth]{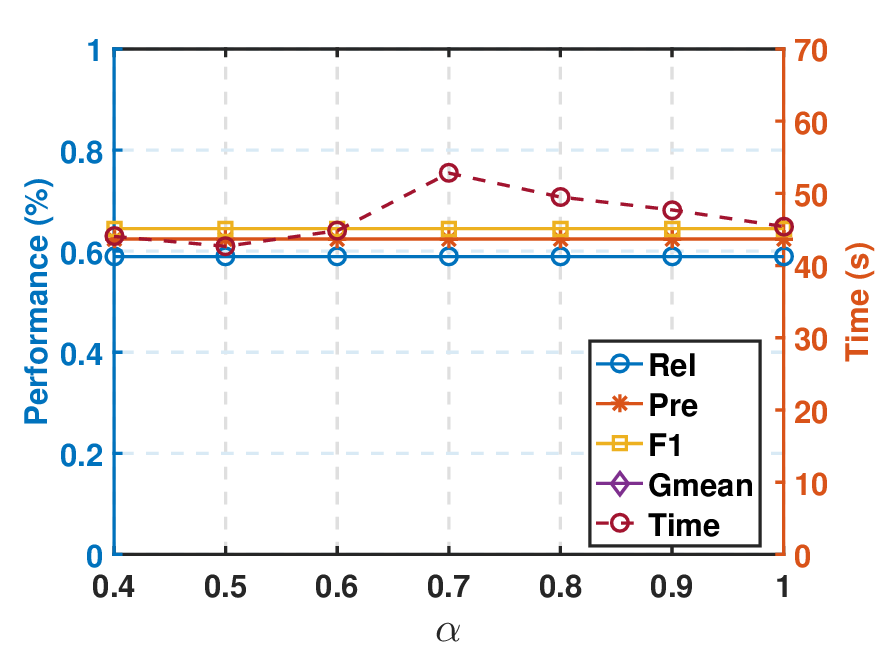}
        \caption{CARLS-M}
    \end{subfigure}
    
    \caption{ Performance of different weights $\alpha$ of pseudo-labeled data on all datasets in class-incremental setting.}
    \label{alph}
\end{figure*}

The detailed results of the parameter sensitivity analysis on all datasets are shown in Figures \ref{RCS}, \ref{CUPL}, and \ref{alph}.

\section{Limitations}
The limitations of the proposed \system mainly include the following three points:

1. Memory Constraints: Retrospect Coreset Selection (RCS) necessitates the storage of historical data, which can lead to substantial memory usage. Efficiently managing this data without compromising performance remains a significant challenge, particularly in long-term deployments.

2. Dependency on Consistent Data Streams: The effectiveness of Stream-based Real-Time Fault Detection (\system) is highly dependent on a continuous and consistent stream of data for ongoing learning and adaptation. Any interruptions or inconsistencies in the data flow can adversely affect the model's performance and its ability to accurately diagnose faults.

3. Scalability to Diverse Industrial Applications: Although \system is designed to be scalable, adapting the framework to different industrial applications—each with varying types of equipment, fault conditions, and operational environments—may require significant customization. This need for extensive adaptation can limit the framework’s generalizability and ease of implementation across diverse industries.

\section{pseudo-code}
The pseudocode of the proposed \system is shown in Algorithm 1.

\begin{algorithm}[H]
\caption{SRTFD Algorithm}\label{alg:alg1}
\begin{algorithmic}
\STATE \textbf{Initialization:} Initialize the network $f_{\theta_0}$ 
\STATE \textbf{Input:} Labeled data $D^t =\{X^t, Y^t\}$ and arrive data $X^t_u$
\STATE \hspace{0.7cm} \textbf{1).} Generated Pseudo-labeled samples by CUPL module
\STATE \hspace{1.1cm} \textbf{1.1} Predict the label $Y^t$ by $f_{\theta_{t-1}}$, compute the probability \\\hspace{1.6cm} for $X^t_u$, and the predict uncertainty by variance $\hat{\sigma}^2$  
\STATE \hspace{1.1cm} \textbf{1.2} Generated the Pseudo-labeled dataset by Eq. (\ref{Pseudo})
\STATE \hspace{0.7cm} \textbf{2).} Select coreset $S^t$ by RSC and GBT modules
\STATE \hspace{1.1cm} \textbf{2.1} Filter redundancy data by Eq.(\ref{filter})
\STATE \hspace{1.1cm} \textbf{2.2} Select coreset  $S^t$ by using Eq. (\ref{core}) and Eq. (\ref{Imbalance})
\STATE \hspace{0.7cm} \textbf{3).} Update model $f_{\theta_t}$ 
\STATE \hspace{1.1cm} \textbf{3.1} updating model $f_{\theta_t}$ by using $S^t, D^t, B^t, \theta_{t-1}$, \\\hspace{1.6cm}Eq. (\ref{Loss_imbalance}) and Eq. (\ref{E_Model_update_FD_1})
\STATE \textbf{Output:} Predict $Y^t_u$ for $X^t_u$ and update model $f_{\theta_t}$  
\end{algorithmic}
\label{alg1}
\end{algorithm}

\end{document}